\documentclass{article}
\usepackage{iclr2027_conference,times}

\usepackage{amsmath}
\usepackage{amssymb}
\usepackage{booktabs}
\usepackage{enumitem}
\usepackage{graphicx}
\usepackage{microtype}
\usepackage{multirow}
\usepackage{tabularx}
\usepackage{algorithm}
\usepackage{algpseudocode}
\usepackage{float}
\usepackage{flafter}
\usepackage{placeins}
\usepackage{needspace}
\usepackage{xcolor}
\usepackage{colortbl}
\usepackage{hyperref}
\usepackage{url}

\newcommand{\method}{\textup{VideoSlate}}
\newcommand{\rextime}{\textsc{ReXTime}}

\newcommand{\IoU}{\operatorname{IoU}}

\title{\method{}: Stateful Commit Control\protect\\
for Temporally Grounded\protect\\
Agentic Video Reasoning}

\author{%
  Shuo Zheng$^{1}$, Yuanjie Dang$^{1}$\thanks{Corresponding author: dangyj@zjut.edu.cn}, Peng Chen$^{1}$,
  Jianjun Xiang$^{1}$\\
  \textbf{Ronghua Liang$^{1}$, Mohammed Bennamoun$^{2}$, Qiuhong Ke$^{3}$}\\
  $^{1}$Zhejiang University of Technology\enspace$^{2}$University of Western Australia\enspace$^{3}$Monash University
}

\iclrfinalcopy

\begin{document}

\maketitle
\lhead{Preprint}
\renewcommand{\headrulewidth}{0.4pt}

\begin{abstract}
Agentic video reasoning systems gather and assess video evidence across stages, selecting relevant spans as visual evidence for answering questions. New candidates and observations prompt reassessment of the current span, but revisions can correct localization errors or discard useful evidence already present in the adopted span. Systems therefore need to connect evidence organization across stages with temporal state control, allowing accumulated information to jointly inform revision assessment while explicit adoption decisions govern how assessment outcomes affect downstream visual evidence. We introduce VideoSlate, a stateful commit-control framework built around shared evidence. A typed evidence blackboard links candidate spans, provenance, construction, and verifier observations, enabling an incumbent-relative certifier to assess revision support from the candidate, current span, and related proposals. Based on certification, the controller authorizes a revision or retains the current span, separating evidence accumulation from state updates and recording each decision and its evidence on the blackboard. Extensive evaluations across video grounding and grounded question-answering benchmarks show that VideoSlate can improve temporal localization through selective revision.
\end{abstract}

\section{Introduction}
\label{sec:introduction}

Agentic video reasoning methods use planning, tool invocation, or role coordination to gather video evidence over multiple steps and form judgments from it \citep{gao2023assistgpt,fan2024videoagent,liu2026videomind}. To use this evidence, the system must account for temporal relations: visually similar content may occur at different times, but only some occurrences satisfy the query's temporal conditions \citep{wang2024videoagentlong}. Subsequent judgments therefore need to relate new observations to the evidence and interpretations already available. If these connections no longer inform later judgments, recorded observations may remain accessible without constraining the interpretation of new evidence. Agentic video reasoning thus needs to organize and maintain evidence so that accumulated temporal context continues to guide the interpretation of new observations and the formation of subsequent judgments.

\begin{figure}[!htbp]
    \centering
    \includegraphics[width=\linewidth]{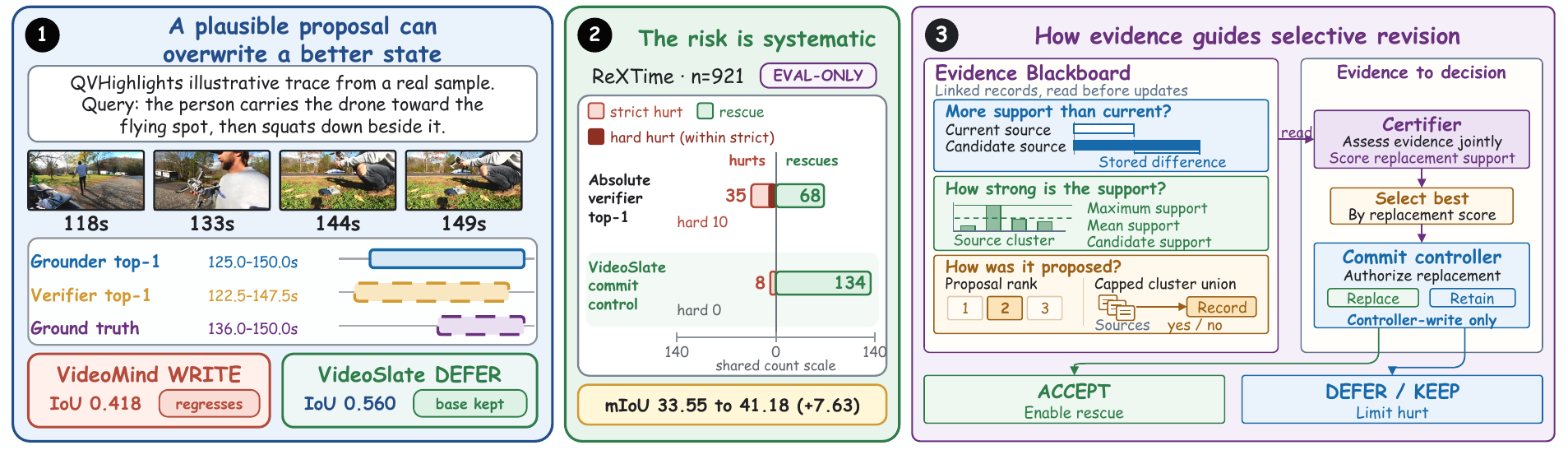}
    \caption{(a) A challenger selected by VideoMind's verifier can overwrite a stronger incumbent;
    deferral preserves it.  (b) Selective commits retain beneficial revisions
    while suppressing harmful writes.  (c) The certifier jointly assesses
    relative support, source support, and proposal context from the evidence
    blackboard. The controller uses the best candidate's replacement score
    to authorize replacement or retain the current result.}
    \label{fig:motivation}
\end{figure}

Memory and retrieval help maintain these connections. VideoAgent uses structured event and object memories with tool queries, while LangRepo organizes descriptions at multiple temporal scales \citep{fan2024videoagent,kahatapitiya2024langrepo}. DrVideo iteratively retrieves and augments a video document, and more recent agents maintain hierarchical or multimodal memories for adaptive tool use and retrieval \citep{ma2025drvideo,yin2026videoarm,yeo2026worldmm}. These systems make prior information available for later reasoning, often through active reuse. However, storing or retrieving evidence does not by itself determine how to compare observations that support different temporal interpretations. The remaining question is what each observation supports and how conflicting support should affect the judgment carried forward.

When later reasoning relies on the currently adopted interval, resolving such disagreements requires determining whether the available evidence is sufficient to change the incumbent and the visual content it supplies downstream. A candidate's match to the query characterizes support for that candidate in isolation; deciding whether to revise the current temporal judgment also requires comparing that support with the evidence for the existing span. VideoMind provides one concrete instance of the former: its verifier scores each candidate independently according to whether it accurately covers the queried event and selects the highest-scoring proposal \citep{liu2026videomind}. Figure~\ref{fig:motivation}(a) illustrates the resulting risk: the verifier prefers a shifted candidate that has less overlap with the reference span than the initial localization. This comparison is further complicated by candidate construction: combined or modified intervals can inherit support assessed for source intervals with different temporal boundaries.

To make information from different reasoning stages jointly inform temporal judgments, we connect candidate generation and evaluation through shared evidence and use explicit state control to coordinate how evaluation results affect subsequent reasoning. Shared evidence preserves links among candidates, observations, and their sources, allowing revision assessment to combine support for the candidate, related proposals, and the incumbent. This evidence jointly informs whether to revise the adopted span, while the resulting recommendation requires a separate commit decision before changing the temporal state used downstream. The aim is to decouple evidence accumulation from state updates, with the former expanding the basis for comparison and the latter controlling which revisions are adopted. Through this division of responsibility, candidate generation, evidence assessment, and result adoption are connected through shared evidence. Information from each stage contributes to temporal judgments, while explicit state-update rules control its effect on the downstream visual context.

Guided by these principles, we introduce \method{}, a stateful commit-control framework. The framework incorporates a typed evidence blackboard that stores the incumbent alongside candidate intervals, linking each candidate to its provenance, construction, and verifier observations. A learned incumbent-relative certifier scores each proposed replacement based on features from these records, combining verifier support for the candidate and incumbent, aggregated support from related proposals, and candidate provenance. A dedicated controller selects the candidate with the highest certifier score and authorizes replacement only when that score meets the acceptance threshold; otherwise, it retains the incumbent. The controller records each decision and its supporting evidence on the blackboard, allowing the returned interval to be traced to the evidence and commitment rule that produced it. Candidate records remain separate from the committed span: tools append candidates and observations, while only the controller can change the interval used downstream.

The contributions are:
\begin{itemize}[leftmargin=*,itemsep=2pt,topsep=3pt]
    \item We introduce \method{}, a framework that models temporal refinement as state updates under asymmetric replacement risk, explicitly decoupling proposal, certification, and commitment to improve localization through beneficial revisions and provide more precise visual evidence downstream.
    \item We design a typed, append-only evidence blackboard that ensures traceable replacement decisions by explicitly linking proposals, provenance, verifier observations, and traces while keeping candidate records separate from the committed span.
    \item 
    We propose an incumbent-relative certifier that learns replacement value directly from pairs, paired with a controller that dynamically commits revisions based on runtime evidence or defaults to the incumbent when support is insufficient.
\end{itemize}

\section{Related Work}
\label{sec:related_work}

\paragraph{Temporally Grounded Video Understanding.}
Temporal grounding localizes query-relevant events; grounded VideoQA also requires evidence-supported answers. TALL/CTRL refines windows \citep{gao2017tall}; ActivityNet Captions supplies localized descriptions \citep{krishna2017dense}. QVHighlights and UMT address moment retrieval and highlight detection \citep{lei2021qvhighlights,liu2022umt}. Unified supervision, image--language tuning, and diffusion further extend grounding \citep{lin2023univtg,liu2024r2tuning,li2023momentdiff}. Timestamp-aware models support localization \citep{huang2024vtimellm,ren2024timechat,huang2024lita,chen2024timemarker}, with extensions using generation, experts, reinforcement learning, and curated data \citep{li2025unitime,yang2025timeexpert,wang2025timer1,zhang2026timelens}. NExT-GQA, \rextime{}, CG-Bench, and ETBench evaluate grounded video reasoning \citep{xiao2024nextgqa,chen2024rextime,chen2024cgbench,liu2024etbench}. We study whether generated candidates should replace the retained span.

\paragraph{Agentic Video Systems and State Control.}
Agentic video systems compose tools \citep{suris2023vipergpt,yang2023mmreact,gao2023assistgpt} and organize evidence through textual repositories or memory \citep{zhang2023llovi,kahatapitiya2024langrepo,wang2024videoagentlong,fan2024videoagent}. SeViLA couples localization and answering \citep{yu2023sevila}. VideoMind passes the verifier's top proposal to the answerer \citep{liu2026videomind}. REVISOR uses initial reasoning to identify a video interval for denser resampling, then combines the new visual evidence with the initial reasoning trace to refine its answer \citep{li2026revisor}. Beyond video, bMAS/LbMAS coordinates agents through shared blackboard messages; its control unit uses the query, accumulated messages, and agent capabilities to select participants for each round \citep{han2025blackboard}. \method{} certifies revisions against the incumbent and separately authorizes changes to the span used downstream.

\paragraph{Candidate Verification and Selective Prediction.}
Verification assesses quality; selective prediction governs reliance on outputs. Verifiers, process supervision, and search evaluate solutions \citep{cobbe2021verifiers,lightman2024letsverify,wang2024llmmcts}; pairwise learning-to-rank estimates preferences \citep{burges2005ranknet}. Selective models trade coverage for risk \citep{geifman2017selective,geifman2019selectivenet}; deferral routes cases to experts \citep{madras2018defer,mozannar2020defer}.  \method{} applies selective prediction to revision, retaining the incumbent when evidence is insufficient and logging the decision.

\section{VideoSlate: Selective Temporal State Revision}
\label{sec:method}

\subsection{Problem Formulation and Overview}

Given a video $V$ of duration $T$ and a query $q$, temporal grounding provides a span whose visual content can support downstream reasoning. Once this span is in use, a proposed revision must be assessed relative to what it would replace. We denote the currently retained span by $b_t=[s_t,e_t]$, the incumbent, and describe one revision step indexed by $t$. A challenger is a candidate proposed to replace this span. The task is to determine whether the available evidence supports a replacement or retaining the incumbent. The resulting span $b_{t+1}$ supplies the visual evidence used downstream.

\begin{figure*}[!htbp]
    \centering
    \includegraphics[width=\textwidth]{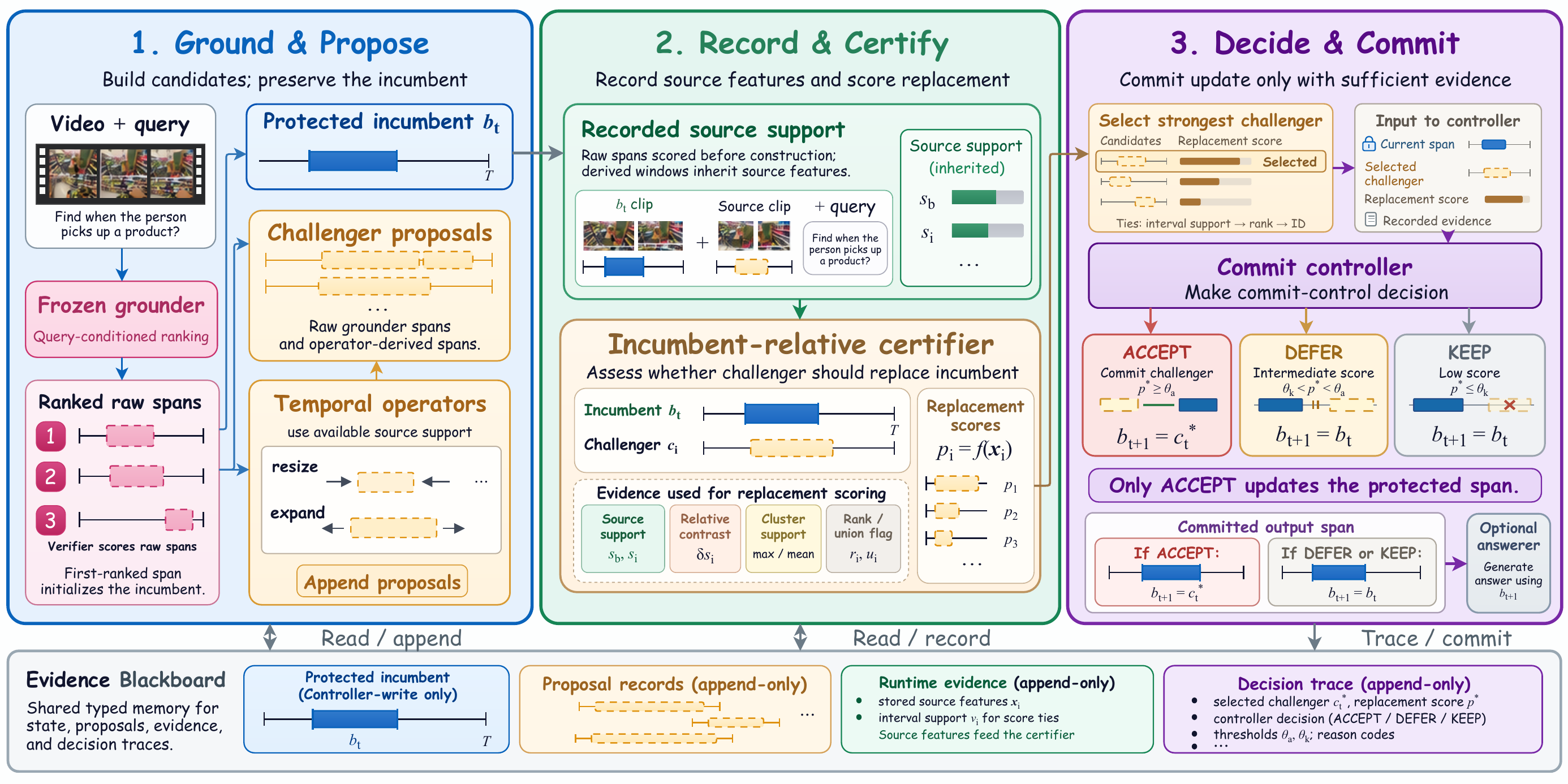}
    \caption{VideoSlate's selective temporal state revision. Ground \& Propose obtains raw spans and their verifier support before constructing additional candidates. Record \& Certify links candidates to their sources and scores replacement using stored source features. Decide \& Commit selects the strongest challenger, using separate interval observations to resolve score ties when available. Only acceptance changes the committed span; both retention actions preserve it with distinct traces. The blackboard records proposals, evidence, and decisions throughout the step. Score bars are schematic.}
    \label{fig:architecture}
\end{figure*}

Making this decision requires keeping the incumbent available as a comparison reference while gathering evidence for alternative spans. We organize this information in a typed evidence blackboard $\mathcal{B}_t$, which stores the incumbent alongside candidate records, verifier observations, and decision traces. Candidate identifiers link each proposed interval to its source proposals and observations. Tools append proposals and observations while the incumbent remains protected; only the controller can replace it. This separation lets evidence accumulate without changing the committed span. Figure~\ref{fig:architecture} organizes this step into Ground \& Propose, Record \& Certify, and Decide \& Commit.

\subsection{Ground \& Propose: Constructing Revision Candidates}
\label{sec:candidate_surface}

A grounder's proposed interval may omit part of the queried event or include unrelated context. Revision therefore needs alternatives that can reduce or extend its temporal coverage. Given $V$ and $q$, the frozen VideoMind grounder \citep{liu2026videomind} produces ranked temporal spans. The primary configuration uses the first-ranked output as the initial incumbent span $b_0$; all ranked outputs, including this span, serve as raw proposals. The verifier first scores these proposals against $q$. These raw proposals and their source-support scores provide the inputs for constructing alternative spans.

Using these inputs, temporal operators construct candidates that preserve, narrow, or broaden the proposed coverage. Direct candidates retain the raw proposal boundaries. Nonempty intersections between the incumbent and a raw proposal restrict coverage to their shared region, while duration edits expand a proposal to include additional context. Union operators combine the incumbent with a raw proposal or combine proposals within a verifier-supported cluster. For cluster construction, the raw proposal with the highest source support serves as the anchor; sufficiently supported proposals join when they overlap the anchor or lie near its center.

These constructions can yield repeated intervals, which we consolidate in the evaluation collection while retaining their source information. Each derived candidate inherits source-support features and retains links to the proposals used to construct it. Such inherited support does not directly verify the boundaries of the resulting interval. Merging duplicate intervals combines these links and takes the maximum of each stored numeric feature, allowing one candidate to carry evidence from several constructions. The resulting raw and derived candidates form the challenger set $\mathcal{C}_t$ for certification, excluding records with the incumbent's boundaries. The incumbent remains available as the fallback output.

\subsection{Record \& Certify: Incumbent-Relative Revision Evidence}
\label{sec:blackboard}
\label{sec:certifier}

Once the candidate set is constructed, certification assesses whether each challenger should replace the incumbent. A candidate's absolute support does not establish improvement over the incumbent. We therefore compare the two spans using the stored source features linked on the blackboard, whose support scores originate from the frozen VideoMind verifier \citep{liu2026videomind}.

To represent replacement evidence, we combine incumbent support as the comparison baseline with candidate and cluster support for the alternative. The stored support difference adds an explicit relative comparison, while rank and a cluster-union indicator encode proposal order and construction type. For challenger $c_i$, these features give the seven-dimensional input
\begin{equation}
\mathbf{x}_i=[s_b,\ r_i,\ s_i,\ s_i^{\max},\ \bar{s}_i,\ u_i,\ \delta s_i]^{\top}.
\label{eq:runtime_features}
\end{equation}
In Equation~\ref{eq:runtime_features}, $s_b$ and $s_i$ are the stored incumbent and candidate support scores. The maximum $s_i^{\max}$ and mean $\bar{s}_i$ summarize the source cluster, which contains a single member for an ordinary proposal. The coordinate $r_i$ stores proposal rank, and $u_i$ indicates whether a capped verifier-cluster union contributed to the record. The stored difference $\delta s_i$ is computed as candidate minus incumbent support before feature inheritance and record merging.

A logistic certifier maps this vector to a replacement score $p_i$ by applying a sigmoid to a learned linear combination with an intercept. Training preferences are defined by overlap improvement relative to the incumbent, so larger scores indicate stronger learned support for replacement. The score encodes this replacement preference; it does not estimate the magnitude of an IoU gain.

To distinguish challengers with equal certifier scores, we retain a separate interval-support value $v_i$. When available, this value is the frozen verifier's yes/no observation of the candidate's own window. Such observations cover a subset of candidate intervals; for a missing observation, $v_i$ falls back to the maximum available stored verifier-support or source-cluster value for that candidate. This observation or fallback provides a tie-breaking input to the controller without changing the source features used to compute $p_i$.

\subsection{Decide \& Commit: Selective State Transition}
\label{sec:decision}

The certifier scores rank challengers by replacement support, but even the highest-scoring candidate may lack enough support to replace the incumbent. The controller therefore first identifies the preferred challenger, then checks whether its score warrants a state change. For a nonempty, scorable challenger set, it selects the preferred challenger using the deterministic lexicographic order
\begin{equation}
i^*=\underset{i:\,c_i\in\mathcal{C}_t}{\arg\max}\big(p_i,\ v_i,\ -r_i,\ \operatorname{id}(c_i)\big).
\label{eq:runtime_argmax}
\end{equation}
Here $i^*$ indexes the selected challenger. The order first maximizes $p_i$, using higher interval support $v_i$ only to break score ties. Remaining ties favor a lower stored rank and, finally, the lexicographically larger stable identifier. Selection uses no reference-dependent information.

The selected score then determines whether the incumbent changes. With frozen thresholds $\theta_{\mathrm{a}}>\theta_{\mathrm{k}}$, Accept applies when $p_{i^*}\geq\theta_{\mathrm{a}}$ and sets $b_{t+1}=c_{i^*}$. Otherwise, the controller retains the incumbent, returning $b_{t+1}=b_t$. It records Defer for a score between the thresholds and Keep for a score at or below $\theta_{\mathrm{k}}$, distinguishing intermediate from low replacement support while returning the same span. An empty or unscorable candidate set also retains the incumbent and records Keep with a fallback reason.

For each such transition, the controller appends a decision trace to the blackboard, linking the returned span to the evidence and rule that produced it. Each trace stores the selected challenger, when available, its score, feature snapshot, thresholds, action, and any fallback reason. For a retained span, these records show whether the controller applied the score thresholds or used a fallback.

\paragraph{Answer generation.}
\label{sec:answerer}
The resulting span $b_{t+1}$ is the localization output of this revision step. For grounded question answering, this span is cropped and passed with $q$ to an answerer outside the replacement policy. Both an accepted challenger and a retained incumbent supply this same downstream interface.

\subsection{Certifier Training and Calibration}
\label{sec:safety}

We learn replacement scores from each incumbent--challenger pair $(b_i,c_i)$ using its overlap change against reference span $g_i$:
\begin{equation}
\Delta_i=\IoU(c_i,g_i)-\IoU(b_i,g_i).
\label{eq:pair_margin}
\end{equation}
Using Equation~\ref{eq:pair_margin}, pairs with $\Delta_i\geq0.05$ receive $y_i=1$, and those with $\Delta_i\leq-0.05$ receive $y_i=0$; ties are excluded from fitting but retained during calibration.

With the grounder and verifier frozen, we fit certifier weights $\mathbf{w}$ and intercept $\beta$ on $N$ labeled pairs:
\begin{equation}
\mathcal{L}(\mathbf{w},\beta)=\sum_{i=1}^{N}\alpha_{y_i}\,\ell_{\mathrm{BCE}}(y_i,p_i)+\frac{\|\mathbf{w}\|_2^2}{2C}.
\label{eq:certifier_objective}
\end{equation}
In Equation~\ref{eq:certifier_objective}, $\ell_{\mathrm{BCE}}$ is binary cross-entropy, $\alpha_y=N/(2N_y)$ balances classes with label counts $N_y$, and $C=1$ controls L2 regularization.

Five video-grouped folds provide out-of-fold (OOF) scores. Since each query returns one span, calibration evaluates outputs after candidate selection. Eligible thresholds improve R@0.5 under harm and retention constraints; selection prioritizes localization gain. Primary OOF score ties use training-candidate IoU and operator-row identifiers; evaluation follows the annotation-free order in Equation~\ref{eq:runtime_argmax}. We select $\theta_{\mathrm{a}}=0.55$ and $\theta_{\mathrm{k}}=0.20$; the latter follows the grid tie convention. The refitted certifier and thresholds are frozen for evaluation.

\section{Experiments}
\label{sec:experiments}

\raggedbottom

We evaluate grounded VideoQA and temporal grounding. Benchmark comparisons assess localization and joint answer--localization quality; controlled comparisons examine how evidence and commitment rules affect revisions.

\begin{table}[t]
    \centering
    \footnotesize
    \setlength{\tabcolsep}{3.2pt}
    \caption{Grounded VideoQA on \rextime{} \citep{chen2024rextime}. Acc@IoU requires a correct answer and temporal IoU $\geq 0.5$. Bold marks the best results; underlining marks the second best.}
    \label{tab:rextime_results}
    \begin{tabular}{lcrrrrr}
        \toprule
        \textbf{Method} & \textbf{Size} & \textbf{R@0.3} &
        \textbf{R@0.5} & \textbf{mIoU} &
        \textbf{Acc} & \textbf{Acc@IoU} \\
        \midrule
        VTimeLLM \citep{huang2024vtimellm} & 7B & 28.84 & 17.41 & 20.14 & 36.16 & -- \\
        TimeChat \citep{ren2024timechat} & 7B & 14.42 & 7.61 & 11.65 & 40.04 & -- \\
        LITA \citep{huang2024lita} & 13B & 29.49 & 16.29 & 21.49 & 34.44 & -- \\
        VTimeLLM (FT) \citep{huang2024vtimellm} & 7B & \underline{43.69} &
        26.13 & 29.92 & 57.58 & 17.13 \\
        TimeChat (FT) \citep{ren2024timechat} & 7B & 40.13 & 21.42 & 26.29 & 49.46 & 10.92 \\
        TOGA \citep{gupta2025toga} & 7B & 29.91 & 19.79 & 25.53 & -- & -- \\
        SynRL \citep{jiang2026synrl} & 4B & 38.80 & 26.20 & 28.90 & -- & -- \\
        Divid \citep{tang2026divid} & 7B & 42.56 & \underline{31.05} &
        \underline{35.78} & -- & \underline{22.26} \\
        VideoMind \citep{liu2026videomind} & 7B & 38.22 & 25.52 & 27.61 &
        \underline{74.59} & 20.20 \\
        \midrule
        \rowcolor{blue!7.5}
        \textbf{VideoSlate (Ours)} & 7B & \textbf{58.85} &
        \textbf{38.98} & \textbf{41.18} & \textbf{75.46} &
        \textbf{31.60} \\
        \bottomrule
    \end{tabular}
\end{table}

\begin{table}[t]
    \centering
    \scriptsize
    \setlength{\tabcolsep}{5pt}
    \caption{Grounded VideoQA on the 3,000-question CG-Bench Mini evaluation
    \citep{chen2024cgbench}.  \emph{long-acc.} is answer accuracy;
    \emph{rec.@IoU} and \emph{acc.@IoU} average localization recall and
    answer-correct hits over IoU thresholds 0.1--0.5.}
    \label{tab:cgbench_audit}
    \begin{tabular}{lcrrrr}
        \toprule
        \textbf{Method} & \textbf{Size} & \textbf{long-acc.} &
        \textbf{mIoU} & \textbf{rec.@IoU} & \textbf{acc.@IoU} \\
        \midrule
        \multicolumn{6}{l}{\textit{Closed-source systems}} \\
        \midrule
        GPT-4o \citep{gpt4o} & -- & \underline{45.2} & 5.62 & 8.30 & 4.38 \\
        Gemini-1.5-Pro \citep{gemini1.5} & -- & 37.2 & 3.95 & 5.81 & 2.53 \\
        Claude-3.5-Sonnet \citep{claude3.5sonnet} & -- & 40.5 & 3.99 & 5.67 & 2.79 \\
        \midrule
        \multicolumn{6}{l}{\textit{Open-source systems}} \\
        \midrule
        Video-LLaVA \citep{videollava} & 7B & 16.2 & 1.13 & 1.96 & 0.59 \\
        VideoLLaMA \citep{videollama} & 7B & 18.4 & 1.21 & 1.87 & 0.84 \\
        VideoChat2 \citep{mvbench} & 7B & 19.3 & 1.28 & 1.98 & 0.94 \\
        ST-LLM \citep{stllm} & 7B & 23.8 & 2.23 & 2.86 & 1.13 \\
        ShareGPT4Video \citep{sharegpt4video} & 16B & 26.7 & 1.85 & 2.65 & 1.01 \\
        Chat-UniVi-v1.5 \citep{chatunivi} & 13B & 25.9 & 2.07 & 2.53 & 1.21 \\
        VILA \citep{vila} & 8B & 28.7 & 1.56 & 2.89 & 1.35 \\
        LongVA \citep{longva} & 7B & 28.7 & 2.94 & 3.86 & 1.78 \\
        LLaVA-OneVision \citep{llavaonevision} & 7B & 31.1 & 1.63 & 1.78 & 1.08 \\
        Video-CCAM \citep{videoccam} & 14B & 29.7 & 2.63 & 3.48 & 1.83 \\
        Kangaroo \citep{kangaroo} & 8B & 30.2 & 2.56 & 2.81 & 1.94 \\
        VITA \citep{vita} & $8\!\times\!7$B & 33.3 & 3.06 & 3.53 & 2.06 \\
        Qwen2-VL \citep{qwen2vl} & 72B & 41.3 & 3.58 & 5.32 & 3.31 \\
        InternVL2 \citep{internvl2} & 78B & 42.2 & 3.91 & 5.05 & 2.64 \\
        VideoMind \citep{liu2026videomind} & 7B & 38.4 & \underline{7.10} & \underline{9.93} & \underline{4.67} \\
        \midrule
        \rowcolor{blue!7.5}
        \textbf{VideoSlate (Ours)} & 7B &
        \textbf{46.6} & \textbf{7.98} & \textbf{12.51} &
        \textbf{7.45} \\
        \bottomrule
    \end{tabular}
\end{table}

\begin{table}[t]
    \centering
    \scriptsize
    \setlength{\tabcolsep}{3.0pt}
    \caption{Localization and grounded-QA performance on the NExT-GQA official test \citep{xiao2024nextgqa}. IoP divides temporal intersection by predicted duration; Acc@GQA requires a correct answer and IoP $\geq0.5$.}
    \label{tab:nextgqa_posthoc}
    \begin{tabular}{>{\raggedright\arraybackslash}p{4.6cm}crrrrrrr}
        \toprule
        \multirow{2}{*}{\textbf{Method}} &
        \multirow{2}{*}{\textbf{Size}} &
        \multicolumn{3}{c}{\textbf{IoU}} &
        \multicolumn{3}{c}{\textbf{IoP}} &
        \multirow{2}{*}{\textbf{Acc@GQA}} \\
        \cmidrule(lr){3-5}\cmidrule(lr){6-8}
        & & R@0.3 & R@0.5 & mIoU & R@0.3 & R@0.5 & mIoP & \\
        \midrule
        FrozenBiLM NG+ \citep{yang2022frozenbilm} & 890M &
        13.50 & 6.10 & 9.60 & 28.50 & 23.70 & 24.20 & 17.50 \\
        SeViLA \citep{yu2023sevila} & 4B &
        29.20 & 13.80 & 21.70 & 34.70 & 22.90 & 29.50 & 16.60 \\
        LangRepo \citep{kahatapitiya2024langrepo} & $8{\times}$7B &
        -- & 12.20 & 18.50 & -- & 28.70 & 31.30 & 17.10 \\
        VideoStreaming \citep{qian2024videostreaming} & 8.3B &
        -- & 13.30 & 19.30 & -- & 31.00 & 32.20 & 17.80 \\
        LLoVi \citep{zhang2023llovi} & 1.8T &
        -- & 15.30 & 20.00 & -- & \textbf{36.90} & 37.30 & 24.30 \\
        HawkEye \citep{wang2024hawkeye} & 7B &
        37.00 & 19.50 & 25.70 & -- & -- & -- & -- \\
        VideoChat-TPO \citep{yan2024videochattpo} & 7B &
        41.20 & 23.40 & 27.70 & 47.50 & 32.80 & 35.60 & 25.50 \\
        VideoMind \citep{liu2026videomind} & 7B &
        \textbf{50.20} & \underline{25.80} & \underline{31.40} & \textbf{56.00} & 35.30 & \underline{39.00} &
        \underline{28.20} \\
        \midrule
        \rowcolor{blue!7.5}
        \textbf{VideoSlate (Ours)} & 7B &
        \underline{49.47} & \textbf{26.54} & \textbf{31.48} & \underline{55.70} &
        \underline{36.81} & \textbf{39.73} & \textbf{30.43} \\
        \bottomrule
    \end{tabular}
\end{table}

\begin{table}[t]
    \centering
    \scriptsize
    \setlength{\tabcolsep}{2.35pt}
    \caption{Video temporal grounding on the Charades-STA official test and
    ActivityNet-Captions.}
    \label{tab:vtg_external}
    \begin{tabular}{lcrrrrrrrr}
        \toprule
        \multirow{2}{*}{\textbf{Method}} &
        \multirow{2}{*}{\textbf{Size}} &
        \multicolumn{4}{c}{\textbf{Charades-STA}} &
        \multicolumn{4}{c}{\textbf{ActivityNet-Captions}} \\
        \cmidrule(lr){3-6}\cmidrule(lr){7-10}
        & & R@0.3 & R@0.5 & R@0.7 & mIoU &
        R@0.3 & R@0.5 & R@0.7 & mIoU \\
        \midrule
        VTimeLLM \citep{huang2024vtimellm} & 7B &
        51.00 & 27.50 & 11.40 & 31.20 & 44.00 & 27.80 & 14.30 & 30.40 \\
        TimeChat \citep{ren2024timechat} & 7B &
        51.50 & 32.20 & 13.40 & -- & -- & -- & -- & -- \\
        Momentor \citep{qian2024momentor} & 7B &
        42.60 & 26.60 & 11.60 & 28.50 & 42.90 & 23.00 & 12.40 & 29.30 \\
        ChatVTG \citep{qu2024chatvtg} & 7B &
        52.70 & 33.00 & 15.90 & 34.90 & 40.70 & 22.50 & 9.40 & 27.20 \\
        E.T. Chat \citep{liu2024etbench} & 4B &
        65.70 & 45.90 & 20.00 & 42.30 & 24.10 & 12.80 & 6.10 & 18.90 \\
        UniTime \citep{li2025unitime} & 7B &
        -- & \underline{59.10} & \underline{31.90} & \textbf{52.20} &
        -- & 22.80 & 14.10 & 27.30 \\
        ED-VTG \citep{pramanick2025edvtg} & -- &
        -- & -- & -- & -- & 52.10 & \underline{33.10} & 16.00 & 35.20 \\
        VideoChat-R1.5 \citep{yan2025videochatr15} & 7B &
        -- & -- & -- & -- & \underline{52.40} & 32.30 &
        \underline{16.80} & \underline{35.50} \\
        VideoMind \citep{liu2026videomind} & 7B &
        \underline{73.50} & \underline{59.10} & 31.20 & 50.20 &
        48.40 & 30.30 & 15.70 & 33.30 \\
        \midrule
        \rowcolor{blue!7.5}
        \textbf{VideoSlate (Ours)} & 7B &
        \textbf{74.81} & \textbf{61.16} & \textbf{33.20} &
        \underline{51.19} & \textbf{55.31} & \textbf{35.95} &
        \textbf{18.00} & \textbf{37.68} \\
        \bottomrule
    \end{tabular}
\end{table}

\subsection{Experimental Settings}

\paragraph{Setup and scope.}
We evaluate \rextime{} validation (921 examples), CG-Bench Mini (3,000 questions), NExT-GQA official test (5,553 queries), Charades-STA official test (3,720 queries), and ActivityNet-Captions (17,031 queries). The first three evaluate grounded QA; the last two isolate localization. The primary \rextime{} configuration uses the frozen VideoMind-7B grounder/verifier. Its offline evaluation replays one state transition per query using frozen candidates and observations. Other benchmarks use dataset-specific policies. In VideoSlate rows, Size denotes the 7B localization backbone; Qwen3-VL-8B \citep{bai2025qwen3vl} answers from the selected temporal context. Published baselines retain their own answering configurations. CG-Bench baseline results follow Table~2 of \citet{liu2026videomind}.

\begin{table}[!t]
    \centering
    \footnotesize
    \setlength{\tabcolsep}{5pt}
    \caption{Module-aligned controls on the \rextime{} replay. Feature and candidate interventions retain other coefficients and thresholds. QA uses VideoMind-7B for controls and Qwen3-VL-8B for the full policy; joint metrics compare systems.}
    \label{tab:rextime_analysis}
    \begin{tabular}{lrrrrr}
        \toprule
        \textbf{Stage and intervention} & \textbf{R@0.3} & \textbf{R@0.5} &
        \textbf{mIoU} & \textbf{Acc} & \textbf{Acc@IoU} \\
        \midrule
        \multicolumn{6}{l}{\textit{Ground \& Propose}} \\
        Cluster-union proposals only & 41.04 & 25.41 & 28.32 & 74.27 & 19.98 \\
        \midrule
        \multicolumn{6}{l}{\textit{Record \& Certify}} \\
        w/o cluster-support terms & 42.13 & 26.60 & 29.46 & 74.48 & 21.17 \\
        w/o union-indicator term & 41.37 & 25.84 & 28.61 & 74.48 & 20.63 \\
        \midrule
        \multicolumn{6}{l}{\textit{Decide \& Commit}} \\
        KEEP only (protected incumbent) & 40.28 & 25.30 & 27.98 & 74.59 & 20.20 \\
        Unconditional verifier top-1 & 46.91 & 28.88 & 33.55 & 73.62 & 22.37 \\
        Fixed verifier-margin gate & 42.78 & 27.90 & 31.36 & 74.05 & 22.26 \\
        \midrule
        \rowcolor{blue!7.5}
        \textbf{VideoSlate (full policy)} & \textbf{58.85} & \textbf{38.98} &
        \textbf{41.18} & \textbf{75.46} & \textbf{31.60} \\
        \bottomrule
    \end{tabular}
\end{table}

\begin{table}[t]
    \raggedright
    \begin{minipage}[t]{0.5\linewidth}
    \vspace{0pt}
    \footnotesize
    \setlength{\tabcolsep}{3pt}
    \caption{Commit allocation on \rextime{} with a 50\% budget and one shared challenger per query. Unselected queries retain their incumbents. Ties use a fixed order; random allocation uses one run. Localization metrics (\%) cover all queries.}
    \label{tab:matched-commit-budget}
    \vspace{3pt}
    \begin{tabularx}{\linewidth}{@{}>{\raggedright\arraybackslash}Xrrr@{}}
        \toprule
        \textbf{Allocation rule} & \textbf{R@0.3} & \textbf{R@0.5} &
        \textbf{mIoU} \\
        \midrule
        Random allocation & 48.97 & 31.49 & 34.22 \\
        Source-support ranking & 48.21 & 30.40 & 33.34 \\
        Source-margin ranking & 50.81 & 32.14 & 35.44 \\
        \midrule
        \rowcolor{blue!7.5}
        \textbf{Certifier-score ranking} & \textbf{51.68} & \textbf{34.20} &
        \textbf{36.37} \\
        \bottomrule
    \end{tabularx}
    \end{minipage}\hfill
    \begin{minipage}[t]{0.47\linewidth}
    \vspace{0pt}
    \footnotesize
    \setlength{\tabcolsep}{2pt}
    \caption{Post-hoc Charades-STA controls on shared frozen candidates. The proposal controls share probability ranking but use a probability threshold or VideoSlate's gate, respectively. Commit counts differ; metrics in \%.}
    \label{tab:charades_commit_controls}
    \vspace{3pt}
    \begin{tabularx}{\linewidth}{@{}>{\raggedright\arraybackslash}Xrrrr@{}}
        \toprule
        \textbf{Commit rule} & \textbf{R@0.3} & \textbf{R@0.5} &
        \textbf{R@0.7} & \textbf{mIoU} \\
        \midrule
        KEEP only & 73.71 & 58.76 & 29.92 & 49.72 \\
        Proposal-prob. gate & 73.82 & 58.92 & 30.16 & 49.83 \\
        Proposal ranking & 74.35 & 60.27 & 32.42 & 50.78 \\
        \midrule
        \rowcolor{blue!7.5}
        \textbf{VideoSlate} & \textbf{74.81} & \textbf{61.16} & \textbf{33.20} &
        \textbf{51.19} \\
        \bottomrule
    \end{tabularx}
    \end{minipage}
\end{table}

\begin{figure}[!t]
    \centering
    \includegraphics[width=\linewidth]{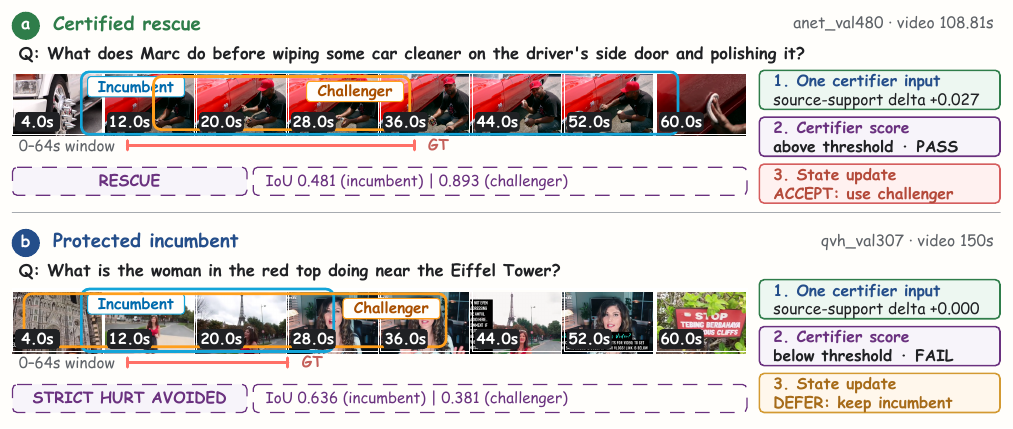}
    \caption{Two outcomes of selective temporal revision. Blue and orange outlines mark the incumbent and challenger; coral reference spans and IoU values are evaluation-only overlays. Source-support difference is one certifier input; the replacement score determines whether the acceptance threshold is met. Top: accepting a challenger yields an R@0.5 rescue. Bottom: deferral preserves the incumbent and avoids a harmful expansion.}
    \label{fig:revision_traces}
\end{figure}

\paragraph{Metrics.}
We report R@0.3, R@0.5, mIoU, and R@0.7 where available. Acc measures answer accuracy; Tables~\ref{tab:rextime_results}--\ref{tab:nextgqa_posthoc} define the benchmark-specific joint metrics. At IoU 0.5, a rescue changes a miss into a hit and a strict hurt changes a hit into a miss; a hard hurt replaces an incumbent at or above 0.5 IoU with an output below 0.3.

\subsection{Comparison with State-of-the-Art Methods}

\paragraph{Grounded VideoQA.}
VideoSlate leads the reported metrics on \rextime{} and CG-Bench (Tables~\ref{tab:rextime_results} and~\ref{tab:cgbench_audit}). On \rextime{}, joint accuracy improves markedly over VideoMind despite similar answer accuracy. Expanded answering windows may cover similar answer-relevant cues even when predicted boundaries differ, while the joint gains highlight VideoSlate's improved temporal grounding of correct responses. CG-Bench provides complementary evidence: among localization hits pooled across thresholds, VideoSlate also answers a larger share correctly than VideoMind.

On NExT-GQA (Table~\ref{tab:nextgqa_posthoc}), VideoSlate achieves the highest reported Acc@GQA, mIoU, and mIoP. Relative to VideoMind, its localization advantage is clearer at the stricter IoU cutoff: coarse recall is slightly lower and mIoU is nearly tied. At the stricter IoP cutoff, recall is close to but slightly below LLoVi, while higher mIoP and Acc@GQA indicate better average temporal overlap and joint QA performance at a comparable localization hit rate.

\paragraph{Video Temporal Grounding.}
Using candidates from the frozen VideoMind grounder, VideoSlate improves all four metrics over VideoMind on Charades-STA and ActivityNet-Captions (Table~\ref{tab:vtg_external}). It leads the reported ActivityNet-Captions metrics and achieves higher recall than UniTime at medium and high IoU thresholds on Charades-STA, although UniTime retains higher mIoU. UniTime uses adaptive frame scaling for boundary prediction \citep{li2025unitime}, while residual candidate-selection errors may limit further mIoU gains in our pipeline.

\subsection{Controlled Comparisons and Ablations}
\label{sec:key_ablations}

\paragraph{Framework ablations and controls.}
Table~\ref{tab:rextime_analysis} presents ablations and controls across VideoSlate's three runtime stages, examining how candidate construction, certification features, and commit rules affect localization. Further candidate controls show that direct proposals retain the gains, whereas including merge candidates adds few rescues but more strict hurts. Useful revisions therefore already exist among the grounder's outputs, and expanding the set does not ensure that they will be selected.

The feature controls separate candidate availability from the evidence needed to adopt a revision. Removing cluster-support terms or the union indicator preserves candidates but lowers scores; with remaining coefficients and the threshold fixed, useful alternatives go unselected. Thus stored evidence affects whether an existing revision opportunity becomes an actual update.

The decision controls further show that preserving correct incumbents can coexist with recovering more errors. The full policy improves both outcomes over unconditional verifier selection, while the tested margin gate falls short in localization. These findings support assessing replacement value from candidate and incumbent evidence before changing the committed span.

\paragraph{Allocation under a fixed commit budget.}
Table~\ref{tab:matched-commit-budget} examines how scoring rules prioritize updates when commit counts are matched. Each query's incumbent and challenger are fixed; only which queries commit changes. Source-support ranking trails random allocation, indicating that candidate support alone is insufficient to prioritize useful replacements. A well-supported candidate may add little when the incumbent is already adequate. Source-margin ranking improves localization, while the certifier gives the best results at 50\%. These results show that jointly using candidate, incumbent, and construction evidence enables more effective update allocation within the shared pool.

\paragraph{Candidate selection and commitment.}
Table~\ref{tab:charades_commit_controls} compares candidate selection and commitment on shared frozen Charades-STA candidates. Proposal ranking replaces VideoSlate's candidate selection with the highest-probability candidate while retaining its commit rule. It outperforms Proposal-prob. gate, which selects the same candidate but uses a probability threshold. VideoSlate further improves all four metrics over Proposal ranking. Since the candidate pool is shared, these gains reflect better use of available alternatives. Higher recall across the reported thresholds and higher mean overlap indicate gains beyond a single overlap cutoff. The commit rule evaluates only the selected challenger, so effective revision depends on both choosing an alternative and deciding whether to adopt it.

\subsection{Visualization}
Figure~\ref{fig:revision_traces} illustrates why a relevant candidate need not improve the current span. In the rescue case, the incumbent already covers the activity, but a tighter challenger removes excess context and improves boundary alignment. Conversely, the deferred merge inherits strong source support while adding only time outside the reference event. This expansion preserves event coverage but reduces temporal precision by including more surrounding context. Retaining the incumbent preserves useful localization despite another relevant interval being available. Both incumbents already cover the reference event, showing that event coverage alone cannot determine whether a revision is useful. The incumbent provides a reference for distinguishing improved boundary precision from expansion that only adds excess context. High source support therefore does not, by itself, establish that a derived candidate is a better replacement. The certifier evaluates candidate and incumbent evidence jointly to assess whether the proposed boundary change warrants replacing the current span.

\FloatBarrier
\flushbottom

\section{Conclusion}
\label{sec:conclusion}

We presented \method{}, a framework for agentic video reasoning that separates proposing temporal revisions from deciding whether to replace the current span. Its evidence blackboard links candidates to their provenance and verifier observations, supplying evidence for incumbent-relative comparison. The certifier assesses replacement value from this evidence, and the controller authorizes supported updates while retaining the incumbent otherwise. Evaluations show localization and joint answer-localization gains, with more corrections and fewer harmful replacements than unconditional verifier selection in the primary controlled evaluation.

\FloatBarrier
\clearpage

\section*{Acknowledgements}

This work was supported in part by the Natural Science Foundation of China under Grant U25A20539, 62432014, 62276237, 62371421, the Zhejiang Provincial Natural Science Foundation of China under Grant LDT23F0202, LDT23F02021F02, the Ten Thousand Talent Program of Zhejiang Province, and the Australian Government through the Australian Research Council's DECRA funding scheme under Grant DE250100030 and Discovery Project funding scheme under Grant DP260100218. Dr Qiuhong Ke is the recipient of an Australian Research Council Discovery Early Career Researcher Award (project number DE250100030) funded by the Australian Government.

\bibliography{references}

@inproceedings{gao2017tall,
  title = {{TALL}: Temporal Activity Localization via Language Query},
  author = {Gao, Jiyang and Sun, Chen and Yang, Zhenheng and Nevatia, Ram},
  booktitle = {Proceedings of the IEEE International Conference on Computer Vision},
  pages = {5267--5275},
  year = {2017},
  url = {https://openaccess.thecvf.com/content_iccv_2017/html/Gao_TALL_Temporal_Activity_ICCV_2017_paper.html}
}

@inproceedings{krishna2017dense,
  title = {Dense-Captioning Events in Videos},
  author = {Krishna, Ranjay and Hata, Kenji and Ren, Frederic and Fei-Fei, Li and Niebles, Juan Carlos},
  booktitle = {Proceedings of the IEEE International Conference on Computer Vision},
  pages = {706--715},
  year = {2017},
  url = {https://openaccess.thecvf.com/content_iccv_2017/html/Krishna_Dense-Captioning_Events_in_ICCV_2017_paper.html}
}

@inproceedings{lei2021qvhighlights,
  title = {Detecting Moments and Highlights in Videos via Natural Language Queries},
  author = {Lei, Jie and Berg, Tamara L. and Bansal, Mohit},
  booktitle = {Advances in Neural Information Processing Systems},
  volume = {34},
  year = {2021},
  url = {https://papers.nips.cc/paper/2021/hash/62e0973455fd26eb03e91d5741a4a3bb-Abstract.html}
}

@inproceedings{lin2023univtg,
  title = {{UniVTG}: Towards Unified Video-Language Temporal Grounding},
  author = {Lin, Kevin Qinghong and Zhang, Pengchuan and Chen, Joya and Pramanick, Shraman and Gao, Difei and Wang, Alex Jinpeng and Yan, Rui and Shou, Mike Zheng},
  booktitle = {Proceedings of the IEEE/CVF International Conference on Computer Vision},
  pages = {2794--2804},
  year = {2023},
  url = {https://openaccess.thecvf.com/content/ICCV2023/html/Lin_UniVTG_Towards_Unified_Video-Language_Temporal_Grounding_ICCV_2023_paper.html}
}

@inproceedings{liu2024r2tuning,
  title = {{$R^2$-Tuning}: Efficient Image-to-Video Transfer Learning for Video Temporal Grounding},
  author = {Liu, Ye and He, Jixuan and Li, Wanhua and Kim, Junsik and Wei, Donglai and Pfister, Hanspeter and Chen, Chang Wen},
  booktitle = {European Conference on Computer Vision},
  year = {2024},
  url = {https://arxiv.org/abs/2404.00801}
}

@inproceedings{huang2024vtimellm,
  title = {{VTimeLLM}: Empower {LLM} to Grasp Video Moments},
  author = {Huang, Bin and Wang, Xin and Chen, Hong and Song, Zihan and Zhu, Wenwu},
  booktitle = {Proceedings of the IEEE/CVF Conference on Computer Vision and Pattern Recognition},
  pages = {14271--14280},
  year = {2024},
  url = {https://openaccess.thecvf.com/content/CVPR2024/html/Huang_VTimeLLM_Empower_LLM_to_Grasp_Video_Moments_CVPR_2024_paper.html}
}

@inproceedings{ren2024timechat,
  title = {{TimeChat}: A Time-Sensitive Multimodal Large Language Model for Long Video Understanding},
  author = {Ren, Shuhuai and Yao, Linli and Li, Shicheng and Sun, Xu and Hou, Lu},
  booktitle = {Proceedings of the IEEE/CVF Conference on Computer Vision and Pattern Recognition},
  pages = {14313--14323},
  year = {2024},
  url = {https://openaccess.thecvf.com/content/CVPR2024/html/Ren_TimeChat_A_Time-sensitive_Multimodal_Large_Language_Model_for_Long_Video_CVPR_2024_paper.html}
}

@inproceedings{qian2024momentor,
  title = {Momentor: Advancing Video Large Language Model with Fine-Grained Temporal Reasoning},
  author = {Qian, Long and Li, Juncheng and Wu, Yu and Ye, Yaobo and Fei, Hao and Chua, Tat-Seng and Zhuang, Yueting and Tang, Siliang},
  booktitle = {Proceedings of the 41st International Conference on Machine Learning},
  series = {Proceedings of Machine Learning Research},
  volume = {235},
  pages = {41340--41356},
  year = {2024},
  url = {https://proceedings.mlr.press/v235/qian24a.html}
}

@inproceedings{qu2024chatvtg,
  title = {{ChatVTG}: Video Temporal Grounding via Chat with Video Dialogue Large Language Models},
  author = {Qu, Mengxue and Chen, Xiaodong and Liu, Wu and Li, Alicia and Zhao, Yao},
  booktitle = {Proceedings of the IEEE/CVF Conference on Computer Vision and Pattern Recognition Workshops},
  pages = {1847--1856},
  year = {2024},
  url = {https://openaccess.thecvf.com/content/CVPR2024W/PVUW/html/Qu_ChatVTG_Video_Temporal_Grounding_via_Chat_with_Video_Dialogue_Large_CVPRW_2024_paper.html}
}

@inproceedings{yang2025timeexpert,
  title = {{TimeExpert}: An Expert-Guided Video {LLM} for Video Temporal Grounding},
  author = {Yang, Zuhao and Yu, Yingchen and Zhao, Yunqing and Lu, Shijian and Bai, Song},
  booktitle = {Proceedings of the IEEE/CVF International Conference on Computer Vision},
  pages = {24286--24296},
  year = {2025},
  url = {https://openaccess.thecvf.com/content/ICCV2025/html/Yang_TimeExpert_An_Expert-Guided_Video_LLM_for_Video_Temporal_Grounding_ICCV_2025_paper.html}
}

@inproceedings{zhang2026timelens,
  title = {{TimeLens}: Rethinking Video Temporal Grounding with Multimodal {LLM}s},
  author = {Zhang, Jun and Wang, Teng and Ge, Yuying and Ge, Yixiao and Li, Xinhao and Wang, Limin},
  booktitle = {Proceedings of the IEEE/CVF Conference on Computer Vision and Pattern Recognition},
  pages = {10419--10429},
  year = {2026},
  url = {https://openaccess.thecvf.com/content/CVPR2026/html/Zhang_TimeLens_Rethinking_Video_Temporal_Grounding_with_Multimodal_LLMs_CVPR_2026_paper.html}
}

@inproceedings{xiao2024nextgqa,
  title = {Can {I} Trust Your Answer? Visually Grounded Video Question Answering},
  author = {Xiao, Junbin and Yao, Angela and Li, Yicong and Chua, Tat-Seng},
  booktitle = {Proceedings of the IEEE/CVF Conference on Computer Vision and Pattern Recognition},
  pages = {13204--13214},
  year = {2024},
  url = {https://openaccess.thecvf.com/content/CVPR2024/html/Xiao_Can_I_Trust_Your_Answer_Visually_Grounded_Video_Question_Answering_CVPR_2024_paper.html}
}

@inproceedings{chen2024rextime,
  title = {{ReXTime}: A Benchmark Suite for Reasoning-Across-Time in Videos},
  author = {Chen, Jr-Jen and Liao, Yu-Chien and Lin, Hsi-Che and Yu, Yu-Chu and Chen, Yen-Chun and Wang, Yu-Chiang Frank},
  booktitle = {Advances in Neural Information Processing Systems Datasets and Benchmarks Track},
  volume = {37},
  year = {2024},
  doi = {10.52202/079017-0900},
  url = {https://proceedings.neurips.cc/paper_files/paper/2024/hash/32683193e1d0e7a5795b073acecb3549-Abstract-Datasets_and_Benchmarks_Track.html}
}

@inproceedings{yu2023sevila,
  title = {Self-Chained Image-Language Model for Video Localization and Question Answering},
  author = {Yu, Shoubin and Cho, Jaemin and Yadav, Prateek and Bansal, Mohit},
  booktitle = {Advances in Neural Information Processing Systems},
  volume = {36},
  year = {2023},
  url = {https://papers.nips.cc/paper_files/paper/2023/hash/f22a9af8dbb348952b08bd58d4734b50-Abstract-Conference.html}
}

@inproceedings{liu2026videomind,
  title = {{VideoMind}: A Chain-of-{LoRA} Agent for Temporal-Grounded Video Reasoning},
  author = {Liu, Ye and Lin, Kevin Qinghong and Chen, Chang Wen and Shou, Mike Zheng},
  booktitle = {International Conference on Learning Representations},
  year = {2026},
  url = {https://proceedings.iclr.cc/paper_files/paper/2026/hash/5e84e4413268b713f0d4a1b23a9dae57-Abstract-Conference.html}
}

@inproceedings{geifman2019selectivenet,
  title = {{SelectiveNet}: A Deep Neural Network with an Integrated Reject Option},
  author = {Geifman, Yonatan and El-Yaniv, Ran},
  booktitle = {Proceedings of the 36th International Conference on Machine Learning},
  pages = {2151--2159},
  volume = {97},
  series = {Proceedings of Machine Learning Research},
  year = {2019},
  url = {https://proceedings.mlr.press/v97/geifman19a.html}
}

@inproceedings{madras2018defer,
  title = {Predict Responsibly: Improving Fairness and Accuracy by Learning to Defer},
  author = {Madras, David and Pitassi, Toniann and Zemel, Richard},
  booktitle = {Advances in Neural Information Processing Systems},
  volume = {31},
  year = {2018},
  url = {https://proceedings.neurips.cc/paper/2018/hash/09d37c08f7b129e96277388757530c72-Abstract.html}
}

@inproceedings{burges2005ranknet,
  title = {Learning to Rank Using Gradient Descent},
  author = {Burges, Chris and Shaked, Tal and Renshaw, Erin and Lazier, Ari and Deeds, Matt and Hamilton, Nicole and Hullender, Greg},
  booktitle = {Proceedings of the 22nd International Conference on Machine Learning},
  year = {2005},
  url = {https://www.microsoft.com/en-us/research/publication/learning-to-rank-using-gradient-descent/}
}

@inproceedings{chen2024cgbench,
  title = {{CG-Bench}: Clue-Grounded Question Answering Benchmark for Long Video Understanding},
  author = {Chen, Guo and Liu, Yicheng and Huang, Yifei and He, Yuping and Pei, Baoqi and Xu, Jilan and Wang, Yali and Lu, Tong and Wang, Limin},
  booktitle = {International Conference on Learning Representations},
  year = {2025},
  url = {https://proceedings.iclr.cc/paper_files/paper/2025/hash/70fa5df8e3300dc30bf19bee44a56155-Abstract-Conference.html}
}

@article{chen2024timemarker,
  title = {{TimeMarker}: A Versatile Video-{LLM} for Long and Short Video Understanding with Superior Temporal Localization Ability},
  author = {Chen, Shimin and Lan, Xiaohan and Yuan, Yitian and Jie, Zequn and Ma, Lin},
  journal = {arXiv preprint arXiv:2411.18211},
  year = {2024},
  url = {https://arxiv.org/abs/2411.18211}
}

@article{cobbe2021verifiers,
  title = {Training Verifiers to Solve Math Word Problems},
  author = {Cobbe, Karl and Kosaraju, Vineet and Bavarian, Mohammad and Chen, Mark and Jun, Heewoo and Kaiser, Lukasz and Plappert, Matthias and Tworek, Jerry and Hilton, Jacob and Nakano, Reiichiro and Hesse, Christopher and Schulman, John},
  journal = {arXiv preprint arXiv:2110.14168},
  year = {2021},
  url = {https://arxiv.org/abs/2110.14168}
}

@inproceedings{fan2024videoagent,
  title = {{VideoAgent}: A Memory-Augmented Multimodal Agent for Video Understanding},
  author = {Fan, Yue and Ma, Xiaojian and Wu, Rujie and Du, Yuntao and Li, Jiaqi and Gao, Zhi and Li, Qing},
  booktitle = {Computer Vision -- ECCV 2024},
  pages = {75--92},
  year = {2024},
  doi = {10.1007/978-3-031-72670-5_5},
  url = {https://doi.org/10.1007/978-3-031-72670-5_5}
}

@article{gao2023assistgpt,
  title = {{AssistGPT}: A General Multi-Modal Assistant That Can Plan, Execute, Inspect, and Learn},
  author = {Gao, Difei and Ji, Lei and Zhou, Luowei and Lin, Kevin Qinghong and Chen, Joya and Fan, Zihan and Shou, Mike Zheng},
  journal = {arXiv preprint arXiv:2306.08640},
  year = {2023},
  url = {https://arxiv.org/abs/2306.08640}
}

@article{han2025blackboard,
  title = {Exploring Advanced {LLM} Multi-Agent Systems Based on Blackboard Architecture},
  author = {Han, Bochen and Zhang, Songmao},
  journal = {arXiv preprint arXiv:2507.01701},
  year = {2025},
  url = {https://arxiv.org/abs/2507.01701}
}

@inproceedings{li2026revisor,
  title = {{REVISOR}: Beyond Textual Reflection, Towards Multimodal Introspective Reasoning in Long-Form Video Understanding},
  author = {Li, Jiaze and Yin, Hao and Tan, Wenhui and Chen, Jingyang and Xu, Boshen and Qu, Yuxun and Chen, Yijing and Ju, Jianzhong and Luo, Zhenbo and Luan, Jian},
  booktitle = {Proceedings of the IEEE/CVF Conference on Computer Vision and Pattern Recognition},
  pages = {5059--5069},
  year = {2026},
  url = {https://openaccess.thecvf.com/content/CVPR2026/html/Li_REVISOR_Beyond_Textual_Reflection_Towards_Multimodal_Introspective_Reasoning_in_Long-Form_CVPR_2026_paper.html}
}

@inproceedings{geifman2017selective,
  title = {Selective Classification for Deep Neural Networks},
  author = {Geifman, Yonatan and El-Yaniv, Ran},
  booktitle = {Advances in Neural Information Processing Systems},
  volume = {30},
  year = {2017},
  url = {https://proceedings.neurips.cc/paper/2017/hash/4a8423d5e91fda00bb7e46540e2b0cf1-Abstract.html}
}

@article{huang2024lita,
  title = {{LITA}: Language Instructed Temporal-Localization Assistant},
  author = {Huang, De-An and Liao, Shijia and Radhakrishnan, Subhashree and Yin, Hongxu and Molchanov, Pavlo and Yu, Zhiding and Kautz, Jan},
  journal = {arXiv preprint arXiv:2403.19046},
  year = {2024},
  url = {https://arxiv.org/abs/2403.19046}
}

@article{kahatapitiya2024langrepo,
  title = {Language Repository for Long Video Understanding},
  author = {Kahatapitiya, Kumara and Ranasinghe, Kanchana and Park, Jongwoo and Ryoo, Michael S.},
  journal = {arXiv preprint arXiv:2403.14622},
  year = {2024},
  url = {https://arxiv.org/abs/2403.14622}
}

@inproceedings{ma2025drvideo,
  title = {{DrVideo}: Document Retrieval Based Long Video Understanding},
  author = {Ma, Ziyu and Gou, Chenhui and Shi, Hengcan and Sun, Bin and Li, Shutao and Rezatofighi, Hamid and Cai, Jianfei},
  booktitle = {Proceedings of the IEEE/CVF Conference on Computer Vision and Pattern Recognition},
  pages = {18936--18946},
  year = {2025},
  url = {https://openaccess.thecvf.com/content/CVPR2025/html/Ma_DrVideo_Document_Retrieval_Based_Long_Video_Understanding_CVPR_2025_paper.html}
}

@inproceedings{yin2026videoarm,
  title = {{VideoARM}: Agentic Reasoning over Hierarchical Memory for Long-Form Video Understanding},
  author = {Yin, Yufei and Meng, Qianke and Chen, Minghao and Ding, Jiajun and Shao, Zhenwei and Yu, Zhou},
  booktitle = {Proceedings of the IEEE/CVF Conference on Computer Vision and Pattern Recognition},
  pages = {24042--24051},
  year = {2026},
  url = {https://openaccess.thecvf.com/content/CVPR2026/html/Yin_VideoARM_Agentic_Reasoning_over_Hierarchical_Memory_for_Long-Form_Video_Understanding_CVPR_2026_paper.html}
}

@inproceedings{yeo2026worldmm,
  title = {{WorldMM}: Dynamic Multimodal Memory Agent for Long Video Reasoning},
  author = {Yeo, Woongyeong and Kim, Kangsan and Yoon, Jaehong and Hwang, Sung Ju},
  booktitle = {Proceedings of the IEEE/CVF Conference on Computer Vision and Pattern Recognition},
  pages = {25599--25609},
  year = {2026},
  url = {https://openaccess.thecvf.com/content/CVPR2026/html/Yeo_WorldMM_Dynamic_Multimodal_Memory_Agent_for_Long_Video_Reasoning_CVPR_2026_paper.html}
}

@inproceedings{li2023momentdiff,
  title = {{MomentDiff}: Generative Video Moment Retrieval from Random to Real},
  author = {Li, Pandeng and Xie, Chen-Wei and Xie, Hongtao and Zhao, Liming and Zhang, Lei and Zheng, Yun and Zhao, Deli and Zhang, Yongdong},
  booktitle = {Advances in Neural Information Processing Systems},
  volume = {36},
  pages = {65948--65966},
  year = {2023},
  url = {https://proceedings.neurips.cc/paper_files/paper/2023/hash/d01bda31bbcd780774ff15b534e03c40-Abstract-Conference.html}
}

@article{li2025unitime,
  title = {Universal Video Temporal Grounding with Generative Multi-Modal Large Language Models},
  author = {Li, Zeqian and Di, Shangzhe and Zhai, Zhonghua and Huang, Weilin and Wang, Yanfeng and Xie, Weidi},
  journal = {arXiv preprint arXiv:2506.18883},
  year = {2025},
  url = {https://arxiv.org/abs/2506.18883}
}

@inproceedings{lightman2024letsverify,
  title = {Let's Verify Step by Step},
  author = {Lightman, Hunter and Kosaraju, Vineet and Burda, Yura and Edwards, Harri and Baker, Bowen and Lee, Teddy and Leike, Jan and Schulman, John and Sutskever, Ilya and Cobbe, Karl},
  booktitle = {International Conference on Learning Representations},
  year = {2024},
  url = {https://proceedings.iclr.cc/paper_files/paper/2024/hash/aca97732e30bcf1303bc22ac3924fd16-Abstract-Conference.html}
}

@inproceedings{liu2022umt,
  title = {{UMT}: Unified Multi-Modal Transformers for Joint Video Moment Retrieval and Highlight Detection},
  author = {Liu, Ye and Li, Siyuan and Wu, Yang and Chen, Chang Wen and Shan, Ying and Qie, Xiaohu},
  booktitle = {Proceedings of the IEEE/CVF Conference on Computer Vision and Pattern Recognition},
  pages = {3042--3051},
  year = {2022},
  url = {https://openaccess.thecvf.com/content/CVPR2022/html/Liu_UMT_Unified_Multi-Modal_Transformers_for_Joint_Video_Moment_Retrieval_and_CVPR_2022_paper.html}
}

@inproceedings{liu2024etbench,
  title = {{E.T. Bench}: Towards Open-Ended Event-Level Video-Language Understanding},
  author = {Liu, Ye and Ma, Zongyang and Qi, Zhongang and Wu, Yang and Shan, Ying and Chen, Chang Wen},
  booktitle = {Advances in Neural Information Processing Systems},
  volume = {37},
  pages = {32076--32110},
  year = {2024},
  url = {https://proceedings.neurips.cc/paper_files/paper/2024/hash/38ace89a980ead30c6be2b46afc1a5bd-Abstract-Datasets_and_Benchmarks_Track.html}
}

@inproceedings{mozannar2020defer,
  title = {Consistent Estimators for Learning to Defer to an Expert},
  author = {Mozannar, Hussein and Sontag, David},
  booktitle = {Proceedings of the 37th International Conference on Machine Learning},
  pages = {7076--7087},
  volume = {119},
  series = {Proceedings of Machine Learning Research},
  year = {2020},
  url = {https://proceedings.mlr.press/v119/mozannar20b.html}
}

@inproceedings{suris2023vipergpt,
  title = {{ViperGPT}: Visual Inference via Python Execution for Reasoning},
  author = {Sur{\'i}s, D{\'i}dac and Menon, Sachit and Vondrick, Carl},
  booktitle = {Proceedings of the IEEE/CVF International Conference on Computer Vision},
  pages = {11888--11898},
  year = {2023},
  url = {https://openaccess.thecvf.com/content/ICCV2023/html/Suris_ViperGPT_Visual_Inference_via_Python_Execution_for_Reasoning_ICCV_2023_paper.html}
}

@article{wang2024llmmcts,
  title = {Towards Self-Improvement of {LLM}s via {MCTS}: Leveraging Stepwise Knowledge with Curriculum Preference Learning},
  author = {Wang, Xiyao and Song, Linfeng and Tian, Ye and Yu, Dian and Peng, Baolin and Mi, Haitao and Huang, Furong and Yu, Dong},
  journal = {arXiv preprint arXiv:2410.06508},
  year = {2024},
  url = {https://arxiv.org/abs/2410.06508}
}

@inproceedings{wang2024videoagentlong,
  title = {{VideoAgent}: Long-Form Video Understanding with Large Language Model as Agent},
  author = {Wang, Xiaohan and Zhang, Yuhui and Zohar, Orr and Yeung-Levy, Serena},
  booktitle = {Computer Vision -- ECCV 2024},
  pages = {58--76},
  year = {2024},
  doi = {10.1007/978-3-031-72989-8_4},
  url = {https://doi.org/10.1007/978-3-031-72989-8_4}
}

@inproceedings{gupta2025toga,
  title = {{TOGA}: Temporally Grounded Open-Ended Video {QA} with Weak Supervision},
  author = {Gupta, Ayush and Roy, Anirban and Chellappa, Rama and Bastian, Nathaniel D. and Velasquez, Alvaro and Jha, Susmit},
  booktitle = {Proceedings of the IEEE/CVF International Conference on Computer Vision},
  pages = {23593--23603},
  year = {2025},
  url = {https://openaccess.thecvf.com/content/ICCV2025/html/Gupta_TOGA_Temporally_Grounded_Open-Ended_Video_QA_with_Weak_Supervision_ICCV_2025_paper.html}
}

@inproceedings{jiang2026synrl,
  title = {Learning Transferable Temporal Primitives for Video Reasoning via Synthetic Videos},
  author = {Jiang, Songtao and Song, Sibo and Zhou, Chenyi and Wang, Yuan and Chen, Ruizhe and Guan, Tongkun and Luo, Ruilin and Zhang, Yan and Tang, Zhihang and Sun, Yuchong and Zhang, Hang and Yang, Zhibo and Bai, Shuai and Lin, Junyang and Liu, Zuozhu},
  booktitle = {Proceedings of the IEEE/CVF Conference on Computer Vision and Pattern Recognition},
  year = {2026},
  url = {https://openaccess.thecvf.com/content/CVPR2026/papers/Jiang_Learning_Transferable_Temporal_Primitives_for_Video_Reasoning_via_Synthetic_Videos_CVPR_2026_paper.pdf}
}

@inproceedings{pramanick2025edvtg,
  title = {Enrich and Detect: Video Temporal Grounding with Multimodal {LLM}s},
  author = {Pramanick, Shraman and Mavroudi, Effrosyni and Song, Yale and Chellappa, Rama and Torresani, Lorenzo and Afouras, Triantafyllos},
  booktitle = {Proceedings of the IEEE/CVF International Conference on Computer Vision},
  pages = {24297--24308},
  year = {2025},
  url = {https://openaccess.thecvf.com/content/ICCV2025/html/Pramanick_Enrich_and_Detect_Video_Temporal_Grounding_with_Multimodal_LLMs_ICCV_2025_paper.html}
}

@inproceedings{tang2026divid,
  title = {{Divid}: Disentangled Spatial-Temporal Modeling within {LLM}s for Temporally Grounded Video Understanding},
  author = {Tang, Yepeng and Wang, Weining and Guo, Longteng and Yue, Tongtian and Wang, Wenxuan and Zhang, Chunjie and Liu, Jing},
  booktitle = {International Conference on Learning Representations},
  year = {2026},
  url = {https://proceedings.iclr.cc/paper_files/paper/2026/hash/1c85c302ece39939c1b334c78f7ee1b8-Abstract-Conference.html}
}

@inproceedings{yan2025videochatr15,
  title = {{VideoChat-R1.5}: Visual Test-Time Scaling to Reinforce Multimodal Reasoning by Iterative Perception},
  author = {Yan, Ziang and Li, Xinhao and He, Yinan and Yue, Zhengrong and Zeng, Xiangyu and Wang, Yali and Qiao, Yu and Wang, Limin and Wang, Yi},
  booktitle = {Advances in Neural Information Processing Systems},
  volume = {38},
  pages = {119152--119184},
  year = {2025},
  url = {https://proceedings.neurips.cc/paper_files/paper/2025/hash/acb94e709f02895fd98b5867f0b184f3-Abstract-Conference.html}
}

@inproceedings{wang2025timer1,
  title = {{Time-R1}: Post-Training Large Vision Language Model for Temporal Video Grounding},
  author = {Wang, Ye and Wang, Ziheng and Xu, Boshen and Du, Yang and Lin, Kejun and Xiao, Zihan and Yue, Zihao and Ju, Jianzhong and Zhang, Liang and Yang, Dingyi and Fang, Xiangnan and He, Zewen and Luo, Zhenbo and Wang, Wenxuan and Lin, Junqi and Luan, Jian and Jin, Qin},
  booktitle = {Advances in Neural Information Processing Systems},
  volume = {38},
  year = {2025},
  url = {https://proceedings.neurips.cc/paper_files/paper/2025/hash/7801b29c93b599b8d0c44138596bdeed-Abstract-Conference.html}
}

@article{yang2023mmreact,
  title = {{MM-ReAct}: Prompting {ChatGPT} for Multimodal Reasoning and Action},
  author = {Yang, Zhengyuan and Li, Linjie and Wang, Jianfeng and Lin, Kevin and Azarnasab, Ehsan and Ahmed, Faisal and Liu, Zicheng and Liu, Ce and Zeng, Michael and Wang, Lijuan},
  journal = {arXiv preprint arXiv:2303.11381},
  year = {2023},
  url = {https://arxiv.org/abs/2303.11381}
}

@inproceedings{zhang2023llovi,
  title = {A Simple {LLM} Framework for Long-Range Video Question-Answering},
  author = {Zhang, Ce and Lu, Taixi and Islam, Md Mohaiminul and Wang, Ziyang and Yu, Shoubin and Bansal, Mohit and Bertasius, Gedas},
  booktitle = {Proceedings of the 2024 Conference on Empirical Methods in Natural Language Processing},
  pages = {21715--21737},
  doi = {10.18653/v1/2024.emnlp-main.1209},
  year = {2024},
  url = {https://aclanthology.org/2024.emnlp-main.1209/}
}

@inproceedings{yang2022frozenbilm,
  title = {Zero-Shot Video Question Answering via Frozen Bidirectional Language Models},
  author = {Yang, Antoine and Miech, Antoine and Sivic, Josef and Laptev, Ivan and Schmid, Cordelia},
  booktitle = {Advances in Neural Information Processing Systems},
  volume = {35},
  pages = {124--141},
  year = {2022},
  url = {https://proceedings.neurips.cc/paper_files/paper/2022/hash/00d1f03b87a401b1c7957e0cc785d0bc-Abstract-Conference.html}
}

@inproceedings{qian2024videostreaming,
  title = {Streaming Long Video Understanding with Large Language Models},
  author = {Qian, Rui and Dong, Xiaoyi and Zhang, Pan and Zang, Yuhang and Ding, Shuangrui and Lin, Dahua and Wang, Jiaqi},
  booktitle = {Advances in Neural Information Processing Systems},
  pages = {119336--119360},
  year = {2024},
  url = {https://proceedings.neurips.cc/paper_files/paper/2024/hash/d7ce06e9293c3d8e6cb3f80b4157f875-Abstract-Conference.html}
}

@article{wang2024hawkeye,
  title = {{HawkEye}: Training Video-Text {LLM}s for Grounding Text in Videos},
  author = {Wang, Yueqian and Meng, Xiaojun and Liang, Jianxin and Wang, Yuxuan and Liu, Qun and Zhao, Dongyan},
  journal = {arXiv preprint arXiv:2403.10228},
  year = {2024},
  url = {https://arxiv.org/abs/2403.10228}
}

@article{yan2024videochattpo,
  title = {Task Preference Optimization: Improving Multimodal Large Language Models with Vision Task Alignment},
  author = {Yan, Ziang and Li, Zhilin and He, Yinan and Wang, Chenting and Li, Kunchang and Li, Xinhao and Zeng, Xiangyu and Wang, Zilei and Wang, Yali and Qiao, Yu and others},
  journal = {arXiv preprint arXiv:2412.19326},
  year = {2024},
  url = {https://arxiv.org/abs/2412.19326}
}

@misc{gpt4o,
  author = {{OpenAI}},
  title = {{GPT-4o} System Card},
  year = {2024},
  url = {https://openai.com/index/gpt-4o-system-card/}
}

@article{gemini1.5,
  author = {Reid, Machel and Savinov, Nikolay and Teplyashin, Denis and Lepikhin, Dmitry and Lillicrap, Timothy and Alayrac, Jean-Baptiste and Soricut, Radu and Lazaridou, Angeliki and Firat, Orhan and Schrittwieser, Julian and others},
  title = {{Gemini 1.5}: Unlocking Multimodal Understanding across Millions of Tokens of Context},
  journal = {arXiv preprint arXiv:2403.05530},
  year = {2024},
  url = {https://arxiv.org/abs/2403.05530v1}
}

@misc{claude3.5sonnet,
  author = {{Anthropic}},
  title = {Introducing {Claude 3.5 Sonnet}},
  year = {2024},
  url = {https://www.anthropic.com/news/claude-3-5-sonnet}
}

@article{videollava,
  author = {Lin, Bin and Zhu, Bin and Ye, Yang and Ning, Munan and Jin, Peng and Yuan, Li},
  title = {{Video-LLaVA}: Learning United Visual Representation by Alignment before Projection},
  journal = {arXiv preprint arXiv:2311.10122},
  year = {2023},
  url = {https://arxiv.org/abs/2311.10122v1}
}

@article{videollama,
  author = {Zhang, Hang and Li, Xin and Bing, Lidong},
  title = {{Video-LLaMA}: An Instruction-Tuned Audio-Visual Language Model for Video Understanding},
  journal = {arXiv preprint arXiv:2306.02858},
  year = {2023},
  url = {https://arxiv.org/abs/2306.02858}
}

@inproceedings{mvbench,
  author = {Li, Kunchang and Wang, Yali and He, Yinan and Li, Yizhuo and Wang, Yi and Liu, Yi and Wang, Zun and Xu, Jilan and Chen, Guo and Luo, Ping and others},
  title = {{MVBench}: A Comprehensive Multi-Modal Video Understanding Benchmark},
  booktitle = {Proceedings of the IEEE/CVF Conference on Computer Vision and Pattern Recognition},
  pages = {22195--22206},
  year = {2024},
  url = {https://openaccess.thecvf.com/content/CVPR2024/html/Li_MVBench_A_Comprehensive_Multi-modal_Video_Understanding_Benchmark_CVPR_2024_paper.html}
}

@inproceedings{stllm,
  author = {Liu, Ruyang and Li, Chen and Tang, Haoran and Ge, Yixiao and Shan, Ying and Li, Ge},
  title = {{ST-LLM}: Large Language Models Are Effective Temporal Learners},
  booktitle = {European Conference on Computer Vision},
  pages = {1--18},
  year = {2024},
  url = {https://doi.org/10.1007/978-3-031-72998-0_1}
}

@inproceedings{sharegpt4video,
  author = {Chen, Lin and Wei, Xilin and Li, Jinsong and Dong, Xiaoyi and Zhang, Pan and Zang, Yuhang and Chen, Zehui and Duan, Haodong and Lin, Bin and Tang, Zhenyu and Yuan, Li and others},
  title = {{ShareGPT4Video}: Improving Video Understanding and Generation with Better Captions},
  booktitle = {Advances in Neural Information Processing Systems},
  pages = {19472--19495},
  year = {2024},
  url = {https://proceedings.neurips.cc/paper_files/paper/2024/hash/22a7476e4fd36818777c47e666f61a41-Abstract-Datasets_and_Benchmarks_Track.html}
}

@article{chatunivi,
  author = {Jin, Peng and Takanobu, Ryuichi and Zhang, Wancai and Cao, Xiaochun and Yuan, Li},
  title = {{Chat-UniVi}: Unified Visual Representation Empowers Large Language Models with Image and Video Understanding},
  journal = {arXiv preprint arXiv:2311.08046},
  year = {2023},
  url = {https://arxiv.org/abs/2311.08046}
}

@inproceedings{vila,
  author = {Lin, Ji and Yin, Hongxu and Ping, Wei and Molchanov, Pavlo and Shoeybi, Mohammad and Han, Song},
  title = {{VILA}: On Pre-Training for Visual Language Models},
  booktitle = {Proceedings of the IEEE/CVF Conference on Computer Vision and Pattern Recognition},
  pages = {26689--26699},
  year = {2024},
  url = {https://openaccess.thecvf.com/content/CVPR2024/html/Lin_VILA_On_Pre-training_for_Visual_Language_Models_CVPR_2024_paper.html}
}

@article{longva,
  author = {Zhang, Peiyuan and Zhang, Kaichen and Li, Bo and Zeng, Guangtao and Yang, Jingkang and Zhang, Yuanhan and Wang, Ziyue and Tan, Haoran and Li, Chunyuan and Liu, Ziwei},
  title = {Long Context Transfer from Language to Vision},
  journal = {arXiv preprint arXiv:2406.16852},
  year = {2024},
  url = {https://arxiv.org/abs/2406.16852}
}

@article{llavaonevision,
  author = {Li, Bo and Zhang, Yuanhan and Guo, Dong and Zhang, Renrui and Li, Feng and Zhang, Hao and Zhang, Kaichen and Zhang, Peiyuan and Li, Yanwei and Liu, Ziwei and others},
  title = {{LLaVA-OneVision}: Easy Visual Task Transfer},
  journal = {arXiv preprint arXiv:2408.03326},
  year = {2024},
  url = {https://arxiv.org/abs/2408.03326}
}

@article{videoccam,
  author = {Fei, Jiajun and Li, Dian and Deng, Zhidong and Wang, Zekun and Liu, Gang and Wang, Hui},
  title = {{Video-CCAM}: Enhancing Video-Language Understanding with Causal Cross-Attention Masks for Short and Long Videos},
  journal = {arXiv preprint arXiv:2408.14023},
  year = {2024},
  url = {https://arxiv.org/abs/2408.14023}
}

@article{kangaroo,
  author = {Liu, Jiajun and Wang, Yibing and Ma, Hanghang and Wu, Xiaoping and Ma, Xiaoqi and Wei, Xiaoming and Jiao, Jianbin and Wu, Enhua and Hu, Jie},
  title = {Kangaroo: A Powerful Video-Language Model Supporting Long-Context Video Input},
  journal = {arXiv preprint arXiv:2408.15542},
  year = {2024},
  url = {https://arxiv.org/abs/2408.15542}
}

@article{vita,
  author = {Fu, Chaoyou and Lin, Haojia and Long, Zuwei and Shen, Yunhang and Dai, Yuhang and Zhao, Meng and Zhang, Yi-Fan and Dong, Shaoqi and Li, Yangze and Wang, Xiong and others},
  title = {{VITA}: Towards Open-Source Interactive Omni Multimodal {LLM}},
  journal = {arXiv preprint arXiv:2408.05211},
  year = {2024},
  url = {https://arxiv.org/abs/2408.05211}
}

@article{qwen2vl,
  author = {Wang, Peng and Bai, Shuai and Tan, Sinan and Wang, Shijie and Fan, Zhihao and Bai, Jinze and Chen, Keqin and Liu, Xuejing and Wang, Jialin and Ge, Wenbin and others},
  title = {{Qwen2-VL}: Enhancing Vision-Language Model's Perception of the World at Any Resolution},
  journal = {arXiv preprint arXiv:2409.12191},
  year = {2024},
  url = {https://arxiv.org/abs/2409.12191}
}

@misc{internvl2,
  author = {{OpenGVLab}},
  title = {{InternVL2}: Better than the Best---Expanding Performance Boundaries of Open-Source Multimodal Models with the Progressive Scaling Strategy},
  year = {2024},
  url = {https://internvl.github.io/blog/2024-07-02-InternVL-2.0/}
}

@article{bai2025qwen3vl,
  title = {{Qwen3-VL} Technical Report},
  author = {Bai, Shuai and Cai, Yuxuan and Chen, Ruizhe and Chen, Keqin and Chen, Xionghui and Cheng, Zesen and Deng, Lianghao and Ding, Wei and Gao, Chang and Ge, Chunjiang and others},
  journal = {arXiv preprint arXiv:2511.21631},
  year = {2025},
  url = {https://arxiv.org/abs/2511.21631}
}
\bibliographystyle{iclr2027_conference}

\clearpage
\appendix
\section*{Appendix}

\makeatletter
\global\@topnum\m@ne
\makeatother

\setcounter{table}{0}
\renewcommand{\thetable}{S\arabic{table}}
\renewcommand{\theHtable}{S\arabic{table}}
\setcounter{figure}{0}
\renewcommand{\thefigure}{S\arabic{figure}}
\renewcommand{\theHfigure}{S\arabic{figure}}
\setcounter{algorithm}{0}
\renewcommand{\thealgorithm}{S\arabic{algorithm}}
\raggedbottom
\urlstyle{same}
\DeclareUrlCommand\path{\urlstyle{same}}
\setcounter{equation}{0}
\renewcommand{\theequation}{S\arabic{equation}}
\renewcommand{\theHequation}{S\arabic{equation}}

\phantomsection\label{sec:supp_start}
In this appendix, we describe how candidates and evidence produce an update, how the certifier is trained, and what each experiment tests.

\begin{enumerate}[label=\textbf{\Alph*.},leftmargin=1.7em,itemsep=2pt,parsep=0pt,topsep=4pt]
\item \hyperref[sec:supp_model]{Model and Implementation}
  \begin{enumerate}[label=\arabic*.,leftmargin=1.6em,nosep]
  \item \hyperref[sec:supp_agent_contract]{Inference Pipeline}
  \item \hyperref[sec:supp_implementation]{Candidate Construction and Runtime Evidence}
  \item \hyperref[sec:supp_training]{Certifier Training and Calibration}
  \end{enumerate}
\item \hyperref[sec:supp_experiments]{Experiments}
  \begin{enumerate}[label=\arabic*.,leftmargin=1.6em,nosep]
  \item \hyperref[sec:supp_additional_results]{Commit-Rule Diagnostics on Charades-STA}
  \item \hyperref[sec:supp_analysis]{Frozen-Policy Diagnostics}
  \item \hyperref[sec:supp_matched_budget]{Matched-Budget Commit Allocation}
  \end{enumerate}
\end{enumerate}

\section{Model and Implementation}
\label{sec:supp_model}

\subsection{Inference Pipeline}
\label{sec:supp_agent_contract}
Given a video $V$ and query $q$, the primary configuration obtains ranked spans from the frozen grounder. The first-ranked span initializes the incumbent $b_t$ on the blackboard; all ranked spans, including this span, serve as raw proposals for candidate construction.

The frozen verifier scores these spans before temporal operators construct additional windows. Duplicate windows are merged with their provenance. Each challenger links to stored source-support features and available interval observations, allowing evidence to accumulate while the incumbent remains fixed.

The frozen certifier uses each challenger's source features relative to the incumbent to compute a replacement score. The controller selects the highest score, resolving ties deterministically with observation support, rank, and identifier. Section~\ref{sec:supp_implementation} specifies this ordering and its missing-observation fallback. The selected score is compared with frozen accept and keep thresholds; intermediate scores defer the update.

Only acceptance replaces $b_t$ with the selected challenger. Both retention actions, an empty collection, or an unscorable collection return $b_t$. For grounded QA, the output interval determines the video crop passed with $q$ to the answerer. Each decision appends the selected challenger and score when available, its action, and any fallback reason to the trace. Algorithm~\ref{alg:replacement} summarizes this transition, and Table~\ref{tab:supp_agent_contract} lists the blackboard records and their update rules. The primary evaluation replays one transition per query from frozen candidate and observation records.

\begin{algorithm}[!htbp]
\caption{One selective \method{} state transition}
\label{alg:replacement}
\begin{algorithmic}[1]
\Require video $V$, query $q$, blackboard $\mathcal{B}_t$ with incumbent $b_t$,
proposal tools $\mathcal{O}$, verifier $\nu$, frozen certifier $f$, thresholds
$\theta_{\mathrm{a}},\theta_{\mathrm{k}}$
\State $(\mathcal{C}_t,\mathcal{B}_t)\gets
\mathrm{BuildCandidatesAndEvidence}(V,q,b_t,\mathcal{O},\nu,\mathcal{B}_t)$
\If{$\mathcal{C}_t=\varnothing$}
    \State $\mathcal{B}_{t+1}\gets\mathrm{AppendFallbackTrace}(\mathcal{B}_t,\mathrm{Keep},b_t)$
    \State \Return $(\mathrm{Keep},\varnothing,\mathcal{B}_{t+1})$
\EndIf
\For{each challenger $c_i\in\mathcal{C}_t$}
    \State $\mathbf{x}_i\gets\mathrm{RuntimeFeatures}(b_t,c_i,\mathcal{B}_t)$;
    $p_i\gets f(\mathbf{x}_i)$
    \If{scoring fails}
        \State $\mathcal{B}_{t+1}\gets\mathrm{AppendFallbackTrace}(\mathcal{B}_t,\mathrm{Keep},b_t)$
        \State \Return $(\mathrm{Keep},\varnothing,\mathcal{B}_{t+1})$
    \EndIf
    \State $v_i\gets\mathrm{IntervalSupportOrFallback}(c_i,\mathcal{B}_t)$;
    $r_i\gets\mathbf{x}_i[2]$
\EndFor
\State $i^*\gets\arg\max_i(p_i,v_i,-r_i,\operatorname{id}(c_i))$
\State $a_t\gets\mathrm{ThreeWayDecision}(p_{i^*},\theta_{\mathrm{a}},
\theta_{\mathrm{k}})$
\State $b_{t+1}\gets\mathrm{GuardedTransition}(b_t,a_t,c_{i^*})$
\State $\mathcal{B}_{t+1}\gets\mathrm{CommitAndRecord}(\mathcal{B}_t,b_{t+1},c_{i^*},
p_{i^*},\mathbf{x}_{i^*},a_t,\theta_{\mathrm{a}},\theta_{\mathrm{k}})$
\State \Return $(a_t,c_{i^*},\mathcal{B}_{t+1})$
\end{algorithmic}
\end{algorithm}

\begin{table}[!htbp]
    \centering
    \footnotesize
    \setlength{\tabcolsep}{3pt}
    \caption{Blackboard records and update rules in VideoSlate.}
    \label{tab:supp_agent_contract}
    \begin{tabularx}{\linewidth}{l>{\raggedright\arraybackslash}X>{\raggedright\arraybackslash}X}
        \toprule
        \textbf{Blackboard record} & \textbf{Stored information} &
        \textbf{Write and update rule} \\
        \midrule
        Incumbent $b_t$ & Currently selected span & Initialized from grounder
        output; replaced only on controller-authorized acceptance \\
        Candidate records $\mathcal{R}_t$ & Boundaries, ranks, provenance,
        and parent links & Appended by proposal construction;
        do not change $b_t$ \\
        Evidence records $\mathcal{E}_t$ & Observations linked to candidates &
        Appended by evidence collection; read during certification and selection \\
        Decision traces $\mathcal{T}_t$ & Selected challenger, score, thresholds,
        action, and fallback reason & Recorded after each decision,
        including retention of $b_t$ \\
        \bottomrule
    \end{tabularx}
\end{table}

\subsection{Candidate Construction and Runtime Evidence}
\label{sec:supp_implementation}
This section details candidate construction and evidence handling in the primary configuration.

\paragraph{Candidate construction.}
\label{sec:supp_surface}
The first-ranked grounder interval initializes the incumbent. Let $a$ be a raw proposal, $w(a)$ its duration, and $m(a)$ its center. Table~\ref{tab:supp_operators} defines deterministic operators that retain raw boundaries, extract overlap, expand context, or combine proposal coverage. Source verifier scores are available before construction; derived windows inherit support and provenance from their source proposals.

\begin{table}[!htbp]
\centering\footnotesize
\caption{Candidate construction operators in the primary configuration.
All times are in seconds; windows are restricted to $[0,T]$.}
\label{tab:supp_operators}
\begin{tabularx}{\linewidth}{>{\raggedright\arraybackslash}p{0.35\linewidth}>{\raggedright\arraybackslash}X}
\toprule
Operator & Construction rule \\
\midrule
Direct proposal & Preserve the boundaries of $a$. \\
Intersection & Use $b_t\cap a$ if its duration is positive. \\
Minimum-duration window & Center at $m(a)$ with width $\max(w(a),\min(64,T))$. \\
Expanded window & Center at $m(a)$ with width $\max(1.25w(a),\min(64,T))$. \\
Capped incumbent--proposal union & Form the smallest interval enclosing $b_t$
and $a$; if longer than $\min(96,T)$, use that width at its midpoint. \\
Capped verifier-cluster union & Form the smallest interval enclosing the
support cluster; if longer than $\min(96,T)$, use that width at the
support-weighted cluster center. \\
\bottomrule
\end{tabularx}
\end{table}

The incumbent $b_t$ remains protected as the fallback; records matching its boundaries are excluded from challenger selection.

The proposal with the highest source support anchors the cluster. Membership requires sufficient source support and either substantial temporal overlap with the anchor or proximity to its center. If no proposal qualifies, the anchor alone is retained. The cluster center is a weighted average of member centers, using source-support weights with a positive lower bound. For windows specified by a center and width, the width is capped by the video duration; windows crossing a video endpoint are shifted inward to preserve that width.

Training retains separate operator rows, including different constructions of the same interval. The evaluation collection instead merges rows with the same query, video, and endpoints rounded to milliseconds. It retains the first interval record in deterministic source order, unions source identifiers, and takes coordinate-wise maxima of numeric source features, including the stored rank coordinate and operator indicators. Thus an evaluation record may combine support and provenance from several constructions.

\paragraph{Runtime evidence.}
\label{sec:supp_score_streams}
We distinguish two score streams. The certifier consumes stored source-support coordinates $s_b$ and $s_i$, source-cluster maximum $s_i^{\max}$ and mean $\bar{s}_i$, rank $r_i$, a cluster-union indicator $u_i$, and stored contrast $\delta s_i$. For a single source proposal, $\delta s_i=s_i-s_b$; derived windows inherit these coordinates before the aggregation described above. An ordinary proposal uses a singleton support cluster. The indicator $u_i$ is one if a capped verifier-cluster union contributes to the record.

Separately, $v_i$ denotes the interval's VideoMind yes/no observation when available. Observations are requested for the incumbent and a prioritized subset of candidates. During replay, a missing candidate observation falls back to the maximum available candidate-support, auxiliary verifier, cluster-maximum, and cluster-mean coordinates, with missing values mapped to zero. This observation or fallback breaks certifier-score ties; it does not overwrite existing source-feature coordinates. Source support and interval observations have different roles and must remain distinct.

Candidate identifiers link interval boundaries, source records, observations, and decision traces. Only the numeric coordinates in Table~\ref{tab:supp_frozen_features} enter the certifier. With $K$ challengers, linear scoring costs $O(7K)$ and selecting a maximum costs $O(K)$; these costs exclude the frozen grounder and verifier.

\paragraph{Selection and fallback.}
The controller maximizes $(p_i,v_i,-r_i,\operatorname{id}_i)$ lexicographically: higher certifier score, higher observation support, lower stored rank, then the lexicographically larger stable identifier. No IoU or reference-dependent field enters this ordering. For the selected challenger $c^*$, a score $p^*\geq\theta_{\mathrm{a}}$ accepts, $p^*\leq\theta_{\mathrm{k}}$ keeps, and an intermediate score defers, with $\theta_{\mathrm{a}}>\theta_{\mathrm{k}}$. Both retention actions return $b_t$; empty or unscorable collections also retain it and record a fallback reason.

\FloatBarrier
\subsection{Certifier Training and Calibration}
\label{sec:supp_training}
\paragraph{Pairwise supervision.}
\label{sec:supp_certifier}
The primary certifier is fitted on training queries whose videos are disjoint from evaluation videos. Evaluation queries supply neither fitting labels nor threshold-selection outcomes.

Each training pair compares a challenger with its incumbent; the incumbent itself is excluded from the challenger pool. The training margin is $\Delta_i=\IoU(c_i,g)-\IoU(b,g)$. Hard regression is checked first; the remaining rows follow Table~\ref{tab:supp_pair_buckets}. Ties are excluded from fitting but retained among candidates scored during OOF selection.

\begin{table}[!htbp]
    \centering
    \footnotesize
    \caption{Train-side pair-label construction.  ``None'' denotes an
    unlabeled tie used for analysis but not binary fitting.}
    \label{tab:supp_pair_buckets}
    \begin{tabular}{lll}
        \toprule
        Bucket & Condition & Label \\
        \midrule
        Hard regression & $\IoU(b,g)\geq0.5$ and $\IoU(c_i,g)<0.3$ & 0 \\
        Strong positive & $\Delta_i\geq0.20$ & 1 \\
        Weak positive & $0.05\leq\Delta_i<0.20$ & 1 \\
        Unlabeled tie & $|\Delta_i|<0.05$ & None \\
        Weak negative & $-0.20<\Delta_i\leq-0.05$ & 0 \\
        Strong negative & $\Delta_i\leq-0.20$ & 0 \\
        \bottomrule
    \end{tabular}
\end{table}

Five folds are formed by shuffling sorted video identifiers with seed 3030 and assigning them cyclically; all queries from a video remain together. Each historical OOF fold fits at most 20,000 labeled rows drawn by shuffling source-ordered rows. One seeded random generator is used sequentially across folds; an independent generator with the same seed draws the final 50,000-row refit. Class balancing is applied in the loss, not by making positive and negative sample counts equal.

\paragraph{Model fitting.}
The model is logistic regression with $C=1$, L2 regularization, an intercept, L-BFGS optimization, tolerance $10^{-4}$, and at most 1,000 iterations. The fold estimator seed is $3030+\text{fold}$ and the final seed is 3030. Features are unstandardized; missing, non-numeric, and non-finite values map to zero. Replay evaluates $p_i=\sigma(\mathbf{w}^{\top}\mathbf{x}_i+\beta)$ directly, clipping the logit to $[-60,60]$ before exponentiation. Table~\ref{tab:supp_frozen_features} defines the ordered input features.

\begin{table}[!htbp]
\centering\footnotesize
\caption{Ordered input features of the primary certifier.
Source-support coordinates and interval observations are distinct.}
\label{tab:supp_frozen_features}
\begin{tabular}{cll}
\toprule
Index & Symbol & Feature \\
\midrule
1 & $s_b$ & Incumbent source support \\
2 & $r_i$ & Stored proposal-rank coordinate \\
3 & $s_i$ & Candidate source support \\
4 & $s_i^{\max}$ & Source-cluster maximum \\
5 & $\bar{s}_i$ & Source-cluster mean \\
6 & $u_i$ & Capped verifier-cluster-union indicator \\
7 & $\delta s_i$ & Stored source-support contrast \\
\bottomrule
\end{tabular}
\end{table}

\paragraph{Threshold calibration.}
\label{sec:supp_historical_calibration}
The historical OOF calibration scores every challenger, then chooses one per training query. Its ties are resolved by training candidate IoU and operator-row identifier after the model score. This training-only use of annotations is part of the historical calibration recipe; the final evaluation instead uses the annotation-free ordering in Section~\ref{sec:supp_implementation}.

The historical sweep considers accept thresholds $\{0.55,0.60,0.65,0.70,0.75,0.80\}$ and keep thresholds $\{0.20,0.25,0.30,0.35,0.40,0.45\}$, with keep below accept. A pair qualifies when:
\begin{enumerate}[leftmargin=*,nosep]
\item output R@0.5 improves over retaining the incumbent;
\item rescues are at least 1.5 times strict hurts;
\item hard hurts are at most one quarter of rescues;
\item at least 50\% of selected challengers in the tie bucket are deferred or kept;
\item at least 40\% of selected challengers in the negative buckets are deferred or kept.
\end{enumerate}
The last two denominators contain only the selected challenger per query in the specified buckets, not every candidate row. Passing pairs are ranked by R@0.5 improvement, rescues minus strict hurts, fewer hard hurts, and final R@0.5. Exact ties retain the first pair in the ascending grid. The selected thresholds are frozen with the final model. Since keep and defer return the same interval, these localization criteria cannot independently optimize their boundary; the keep threshold follows the tie convention.

\FloatBarrier
\section{Experiments}
\label{sec:supp_experiments}

\subsection{Commit-Rule Diagnostics on Charades-STA}
\label{sec:supp_additional_results}

\begin{table}[!htbp]
\centering\footnotesize
\setlength{\tabcolsep}{3pt}
\caption{Charades-STA commit controls on shared frozen candidates. Random commit matches the duration rule's selected challengers and commit count; it averages seeds 3030--3032, with event counts rounded to integers. Other commit counts differ. R/S/H denote rescue, strict hurt, and hard hurt.}
\label{tab:charades_controls}
\begin{tabularx}{\linewidth}{Xrrrrrr}
\toprule
Commit rule & R@0.3 & R@0.5 & R@0.7 & mIoU & Commits & R/S/H \\
\midrule
Retain incumbent & 73.71 & 58.76 & 29.92 & 49.72 & 0 & 0/0/0 \\
Proposal probability & 73.82 & 58.92 & 30.16 & 49.83 & 382 & 21/15/0 \\
Proposal ranking & 74.35 & 60.27 & 32.42 & 50.78 & 1,494 & 105/49/0 \\
Verifier maximum & 75.54 & 60.05 & 31.48 & 50.94 & 2,347 & 158/110/5 \\
Random commit (mean) & 74.96 & 60.04 & 32.24 & 50.79 & 2,176 & 143/96/1 \\
Unconditional commit & \textbf{75.67} & 60.78 & \textbf{33.58} & \textbf{51.57} & 3,720 & 242/167/2 \\
\rowcolor{blue!7.5}
Duration rule & 74.81 & \textbf{61.16} & 33.20 & 51.19 & 2,176 & 176/87/1 \\
\bottomrule
\end{tabularx}
\end{table}

Table~\ref{tab:charades_controls} distinguishes candidate selection, update allocation, and commit frequency on shared Charades-STA candidates. The duration rule selects the challenger closest to the training duration prior and commits only when it is closer than the incumbent. Selecting instead by proposal probability with the same gate lowers all four localization metrics and reduces commits. The duration rule gains more rescues, at the cost of more strict hurts. Random commit uses the same selected challengers and number of replacements as the duration rule but allocates them uniformly across queries. Relative to its mean, the duration rule recovers more errors and discards fewer correct incumbents, improving R@0.5, R@0.7, and mIoU despite slightly lower R@0.3. Unconditional commit instead removes only the incumbent check. Its additional replacements cause more strict hurts than rescues, lowering R@0.5, but improve R@0.3, R@0.7, and mIoU. Better allocation at matched coverage thus coexists with some missed refinements when the rule limits coverage. Verifier maximum changes both candidate ranking and acceptance, so it remains a separate control.

\FloatBarrier
\Needspace{4\baselineskip}
\subsection{Frozen-Policy Diagnostics}
\label{sec:supp_analysis}
\label{sec:supp_rextime_analysis}

\begin{table}[!htbp]
    \centering
    \small
    \setlength{\tabcolsep}{4.2pt}
    \caption{Decision-rule controls on \rextime{} with shared frozen
    candidates and aligned deterministic ties. Scalar gates use interval
    observations; the primary policy uses the pairwise replacement score.}
    \label{tab:supp_rextime_decision_ablation}
    \begin{tabularx}{\linewidth}{XXrrr}
        \toprule
        \textbf{Output policy} & \textbf{Commit rule} & \textbf{R@0.3} &
        \textbf{R@0.5} & \textbf{mIoU} \\
        \midrule
        Protected incumbent & Never replace & 40.28 & 25.30 & 27.98 \\
        Absolute score margin 0.10 &
        $v(c)\!\geq\!0.35,\ v(c)-v(b)\!\geq\!0.10$ &
        42.45 & 27.04 & 30.00 \\
        Absolute score margin 0.05 &
        $v(c)\!\geq\!0.35,\ v(c)-v(b)\!\geq\!0.05$ &
        42.78 & 27.90 & 31.36 \\
        Absolute verifier top-1 & Unconditional $\arg\max_c v(c)$ &
        46.91 & 28.88 & 33.55 \\
        \rowcolor{blue!7.5}
        \textbf{Primary frozen policy} &
        Pairwise score + fixed threshold &
        \textbf{58.85} & \textbf{38.98} & \textbf{41.18} \\
        \bottomrule
    \end{tabularx}
\end{table}
\FloatBarrier

\paragraph{Decision rules.}
Table~\ref{tab:supp_rextime_decision_ablation} compares fixed verifier rules with the learned policy for deciding whether to replace the incumbent. All controls in this subsection reuse the primary model and records, changing only decision rules, score terms, or candidate families without refitting or recalibrating other parameters. In this comparison, the fixed rules use interval observations $v$, while the learned policy draws on source features for both the candidate and incumbent. The verifier rules improved localization over keeping the incumbent, but fell short of the learned policy. Requiring a larger score margin reduced replacements without narrowing this gap. These fixed cutoffs therefore appear less effective at identifying useful replacements.

\begin{table}[!htbp]
    \centering
    \small
    \setlength{\tabcolsep}{3.8pt}
    \caption{Frozen feature-contribution interventions on the \rextime{} evaluation.
    Remaining coefficients and thresholds stay fixed; removing the replacement
    score returns the incumbent. These are score interventions without refitting.}
    \label{tab:supp_rextime_evidence_ablation}
    \begin{tabular}{lcccrrr}
        \toprule
        \textbf{Configuration} & \textbf{Replacement score} & \textbf{Consensus} &
        \textbf{Provenance} & \textbf{R@0.3} & \textbf{R@0.5} &
        \textbf{mIoU} \\
        \midrule
        No replacement score & -- & -- & -- & 40.28 & 25.30 & 27.98 \\
        No cluster terms & $\checkmark$ & -- & $\checkmark$ &
        42.13 & 26.60 & 29.46 \\
        No union-indicator term & $\checkmark$ & $\checkmark$ & -- &
        41.37 & 25.84 & 28.61 \\
        \midrule
        \textbf{Primary frozen policy} & $\checkmark$ & $\checkmark$ & $\checkmark$ &
        \textbf{58.85} & \textbf{38.98} & \textbf{41.18} \\
        \bottomrule
    \end{tabular}
\end{table}
\FloatBarrier

\paragraph{Feature contributions.}
\label{sec:supp_coordinate_ablation}
Table~\ref{tab:supp_rextime_evidence_ablation} examines how the certifier uses evidence to decide whether to replace the incumbent. Removing the cluster support terms or the union indicator takes away positive contributions to its replacement score. With the acceptance threshold fixed, these removals sharply reduced updates and brought localization close to the no-replacement baseline. Useful alternatives remained available, but the reduced scores left many of these revision opportunities unused. This illustrates the certifier's role in connecting candidate evidence to actual updates: generating a useful candidate does not by itself ensure that the policy will adopt it. At the same time, some features encode overlapping information, so deleting a score term does not necessarily remove that information; combining redundant support terms into an equivalent score preserved every decision. The results therefore describe how the fitted score and its threshold work together.

\begin{table}[!htbp]
    \centering
    \footnotesize
    \setlength{\tabcolsep}{4pt}
    \caption{Candidate-family interventions with fixed coefficients and thresholds.
    Provenance-based families may overlap. A/D/K count accept/defer/keep actions; Strict and Hard count the corresponding hurts.}
    \label{tab:supp_rextime_candidate_ablation}
    \begin{tabular}{lrrrcrrr}
        \toprule
        \textbf{Candidate configuration} & \textbf{R@0.3} & \textbf{R@0.5} &
        \textbf{mIoU} & \textbf{A/D/K} & \textbf{Rescue} &
        \textbf{Strict} & \textbf{Hard} \\
        \midrule
        Primary frozen policy & 58.85 & 38.98 & 41.18 & 881/40/0 & 134 & 8 & 0 \\
        Remove direct & 41.48 & 25.52 & 28.51 &
        211/710/0 & 14 & 12 & 1 \\
        Direct only & \textbf{59.50} &
        \textbf{39.41} & 41.36 & 798/123/0 & 132 & \textbf{2} & \textbf{0} \\
        Remove merge candidates & \textbf{59.50} & \textbf{39.41} &
        \textbf{41.40} & 821/100/0 & 132 & \textbf{2} & \textbf{0} \\
        Only merge candidates & 41.04 & 25.41 & 28.32 & 171/666/84 &
        11 & 10 & 1 \\
        \bottomrule
    \end{tabular}
\end{table}

\paragraph{Candidate families.}
Table~\ref{tab:supp_rextime_candidate_ablation} compares the fixed policy with different sets of candidate intervals. Direct candidates retain the original proposal boundaries; merge candidates are formed by taking interval unions. A candidate can belong to both groups if deduplication combines records from both sources. The controls remove or retain each group while always allowing the incumbent to remain unchanged. Most of the benefit came from direct proposals: keeping only these candidates preserved nearly all rescues and reduced strict hurts, while removing merge candidates slightly improved localization. By comparison, the full policy gained a few more rescues at the cost of more harmful replacements, and merge candidates alone provided little net benefit. These results show that adding candidates does not necessarily improve decisions under the frozen selector.

\subsection{Matched-Budget Commit Allocation}
\label{sec:supp_matched_budget}

\begin{table}[!htbp]
    \centering
    \footnotesize
    \setlength{\tabcolsep}{4pt}
    \caption{Original budget grid on the full \rextime{} replay.
    Random (mean) averages evaluation results over three random seeds
    (3030--3032); event counts are rounded to the nearest integer.
    All orderings coincide at 100\%. Localization metrics are percentages;
    Rescue, Strict, and Hard are event counts.}
    \label{tab:supp_matched_budget}
    \begin{tabular}{llrrrrrr}
        \toprule
        Budget & Gate & R@0.3 & R@0.5 & mIoU & Rescue & Strict & Hard \\
        \midrule
        25\% & Candidate confidence & 43.65 & 27.47 & 30.16 & 22 & 2 & 0 \\
        25\% & Verifier margin & 45.93 & 29.32 & 32.24 & 38 & 1 & 0 \\
        25\% & Learned certifier & 46.58 & 29.21 & 32.42 & 37 & 1 & 0 \\
        25\% & Random (mean) & 44.63 & 28.52 & 31.14 & 32 & 3 & 0 \\
        \midrule
        50\% & Candidate confidence & 48.21 & 30.40 & 33.34 & 53 & 6 & 0 \\
        50\% & Verifier margin & 50.81 & 32.14 & 35.44 & 65 & 2 & 0 \\
        50\% & Learned certifier & 51.68 & 34.20 & 36.37 & 83 & 1 & 0 \\
        50\% & Random (mean) & 49.26 & 31.85 & 34.37 & 66 & 6 & 0 \\
        \midrule
        75\% & Candidate confidence & 53.20 & 33.88 & 36.81 & 89 & 10 & 0 \\
        75\% & Verifier margin & 55.16 & 36.16 & 38.94 & 108 & 8 & 0 \\
        75\% & Learned certifier & 56.03 & 37.79 & 39.49 & 119 & 4 & 0 \\
        75\% & Random (mean) & 54.40 & 35.61 & 37.91 & 103 & 8 & 0 \\
        \midrule
        100\% & All orderings & 58.96 & 39.20 & 41.26 & 138 & 10 & 0 \\
        \bottomrule
    \end{tabular}
\end{table}
\FloatBarrier

\paragraph{Results across commit budgets.}
Table~\ref{tab:supp_matched_budget} further compares how different rankings allocate updates under the same commit budget. We fix one challenger per query from the primary model and vary only which queries commit, while evaluating all queries. Rankings use candidate source support, its difference from incumbent support, or the learned replacement score, rather than the separate interval observations. Ties follow a fixed ordering, and allocations were frozen before joining annotations; these offline quotas compare ordering rather than training-calibrated thresholds. Random allocation spends the same budget on the same challengers without using these scores to prioritize queries. At the partial budgets, random allocation achieved higher R@0.3, R@0.5, and mIoU than candidate-confidence ranking, suggesting that higher candidate support does not always imply a more useful replacement. The learned ordering improved on this random reference in both rescues and strict hurts, indicating that its gains came from assigning the same number of updates more effectively. This balance also favored the learned ordering over the other rankings at 50\% and 75\%, although verifier margin retained a slight R@0.5 advantage at 25\%. At 100\%, every query received its fixed challenger, so all rankings coincided: ordering only matters while the budget requires choosing which updates to make. This all-commit control differs from the primary threshold policy, which can retain an incumbent even when a challenger is available.

\begin{table}[!htbp]
    \centering
    \footnotesize
    \setlength{\tabcolsep}{4pt}
    \caption{Learned minus comparator at 50\% budget: effect [95\%
    descriptive paired video-bootstrap interval].  Localization differences
    are percentage points. The random comparator uses one fixed ordering
    (seed 3030).}
    \label{tab:supp_budget_paired}
    \begin{tabular}{lrrr}
        \toprule
        Comparator & $\Delta$R@0.3 & $\Delta$R@0.5 & $\Delta$mIoU \\
        \midrule
        Random (single run) & $+2.71\,[+0.78,+4.50]$ & $+2.71\,[+0.76,+4.35]$ & $+2.15\,[+1.30,+2.93]$ \\
        Candidate confidence & $+3.47\,[+1.09,+5.54]$ & $+3.80\,[+1.53,+5.71]$ & $+3.03\,[+1.93,+4.00]$ \\
        Verifier margin & $+0.87\,[-0.54,+1.99]$ & $+2.06\,[+0.67,+3.09]$ & $+0.93\,[+0.20,+1.49]$ \\
        \bottomrule
    \end{tabular}
\end{table}

\paragraph{Paired comparisons.}
Table~\ref{tab:supp_budget_paired} assesses how the main-budget differences vary with the sampled videos. Each bootstrap draw resamples whole videos and reapplies equal commit quotas, with models, rankings, and tie rules fixed. The R@0.5 and mIoU difference intervals were positive against all three comparators, whereas the R@0.3 interval against verifier margin crossed zero. Thus the evidence is clearer for gains at the stricter cutoff and in mean overlap. Many verifier margins are zero, so this comparison also depends on the fixed tie ordering; alternative allocations were not tested. The intervals do not account for prior inspection of the evaluation set or adjust for multiple comparisons. The companion harm analysis is less conclusive: the strict-hurt difference interval against verifier margin included zero, and even committing every challenger produced no hard hurt. The evidence therefore supports localization gains more clearly than improved rejection of harmful replacements.

\end{document}